\PassOptionsToPackage{table,dvipsnames}{xcolor}
\documentclass{article}
\usepackage{iclr2027_conference,times}
\usepackage[T1]{fontenc}
\usepackage{amsmath,amssymb,graphicx,booktabs}
\usepackage{algorithm}
\usepackage{algpseudocode}
\usepackage{wrapfig}
\usepackage{needspace}
\usepackage{afterpage}
\usepackage{xcolor}
\usepackage{tabularx}
\usepackage{xspace}

\definecolor{darkred}{HTML}{8B0000}

\definecolor{mydarkred}{HTML}{8B1A1A}
\definecolor{myblue}{HTML}{1F4E79}
\usepackage{hyperref}
\usepackage{url}
\makeatletter
\AtBeginDocument{%
  \let\PW@nat@hyper\NAT@hyper@
  \renewcommand{\NAT@hyper@}[1]{\mbox{\PW@nat@hyper{#1}}}%
}
\newcommand{\pageSafeCitep}[1]{\citep{#1}}
\makeatother
\newcommand{\baseparams}{\theta_{0}}
\newcommand{\taskparams}[1]{\theta_{#1}}
\newcommand{\taskdelta}[1]{\Delta_{#1}}
\newcommand{\mergedparams}{\theta_{\mathrm{merge}}}
\newcommand{\taskset}{\mathcal{T}}
\newcommand{\policy}{\pi}
\newcommand{\policyweave}{\textsc{PolicyWeave}\xspace}

\definecolor{policyweaveblue}{rgb}{0.90,0.96,1.00}

\graphicspath{{figures/}}

\title{Train Together or Merge Later? Unifying VLA Experts via a Shared Action Interface}
\author{\parbox{0.96\textwidth}{\centering\normalfont
Zhizhen Zhang\textsuperscript{1}\quad
Yuxia Fu\textsuperscript{1}\quad
Zijian Wang\textsuperscript{1}\quad
Zi Huang\textsuperscript{1}\quad
Yadan Luo\textsuperscript{1}\\[6pt]
\textsuperscript{1}UQMM Lab, The University of Queensland\\[4pt]
{\small\texttt{\{zhizhen.zhang, yuxia.fu, zijian.wang, helen.huang, y.luo\}@uq.edu.au}}
}}
\iclrfinalcopy
\hypersetup{colorlinks=true,
  linkcolor=mydarkred,
  urlcolor=RoyalBlue,
  citecolor=myblue,
  pdftitle={Train Together or Merge Later? Unifying VLA Experts via a Shared Action Interface},
  pdfauthor={Zhizhen Zhang, Yuxia Fu, Zijian Wang, Zi Huang, Yadan Luo}}
\begin{document}
\maketitle
\lhead{}
\renewcommand{\headrulewidth}{0pt}
\begin{abstract}
Co-training offers a straightforward way to build a multi-task vision-language-action (VLA) policy, 
but can fall short of the performance achieved by training each task independently. 
The challenge is to retain these task-specific gains in a multi-task policy without joint post-training.
Combining independently trained experts through model merging is a natural approach, 
yet strong individual experts do not necessarily yield a strong merged policy. We identify one
source of this \textit{incompatibility}:
task-specific changes to the \emph{action interface}, comprising action
normalization and the action encoder and decoder.
We propose \policyweave, combining merge-compatible post-training with
context-guided sparse merging. During post-training, all experts retain the
common base policy's action interface, while task adaptation is restricted to
LoRA updates in the hidden layers of the action model.
This makes the experts more compatible with existing model merging methods.
However, merging all experts can still introduce interference from unrelated
tasks at deployment. \policyweave scores each expert's LoRA updates using the
initial visual-language context, determines the expert set through
leave-one-layer-out ranking stability, and forms a sparse weighted merge of
the selected updates that remains fixed for current task.
We evaluate \policyweave with GR00T N1.5 on 18 RoboCasa365 tasks,
using only 10\% of the target-task demonstrations for supervised fine-tuning (SFT).
Preserving the shared action interface raises the average success rate
across four static merging methods from 17.0\% to 52.8\%.
\policyweave achieves 64.7\% success with these SFT experts and 74.1\% after
task-specific reinforcement learning (RL), compared with 60.7\% for joint RL.
Further evaluations on LIBERO-10 and an AgileX Piper arm support the deployment
of independently learned skills in long-horizon and real-world manipulation.
Project page: \url{https://policyweave.github.io/}.

\end{abstract}
\newlength{\introSavedTextFloatSep}
\setlength{\introSavedTextFloatSep}{\textfloatsep}
\setlength{\textfloatsep}{12pt}
\raggedbottom
\begingroup
\setlength{\parskip}{2pt}
\setlength{\textfloatsep}{12pt plus 1pt minus 1pt}
\section{Introduction}
\label{sec:introduction}

Vision-language-action (VLA) models learn broadly applicable robot control
capabilities through pretraining on diverse robot data
\citep{brohan2023rt2,ghosh2024octo,kim2025openvla,black2025pi0,bjorck2025groot}.
During downstream post-training, supervised fine-tuning (SFT) adapts VLA
policies using task demonstrations, and reinforcement
learning (RL) can further improve performance through interaction
\citep{li2026simplevlarl,chen2025pirl}.
At either stage, co-training offers a straightforward way to obtain a single policy for multiple tasks. However, balancing competing task objectives can limit performance on individual tasks \citep{yu2020pcgrad}.
Alternatively, task-specific adaptation optimizes each expert independently
from the same pretrained model, but leaves the resulting capabilities in
separate policies
(Figure~\ref{fig:closest-method-comparison}, left).
We therefore ask: \textit{can we retain the gains from task-specific training in a multi-task policy without joint post-training?}

Model merging \citep{ilharco2023taskarithmetic,yadav2023ties} combines parameter updates from independently trained experts without joint retraining and has shown promise in vision and language models.
For VLAs, however, strong individual performance does not guarantee that an
expert's capabilities will survive merging \citep{fu2026mergevla}.
We identify one source of this incompatibility in the \emph{action interface}:
the normalization statistics and action encoder and decoder that map between
physical actions and model representations. During task-specific post-training,
changes to this interface can be compensated for by changes to the expert's
hidden-layer parameters. Merging both can break this compensation and alter
the resulting actions. A simple linear example shows that parameter averaging
can change the output even when the original experts produce identical actions
(Appendix~\ref{app:shared-interface-analysis}). This motivates maintaining
compatibility while experts learn their task-specific capabilities.
To this end, we introduce \policyweave, which
keeps the base policy's action interface fixed while learning task-specific
LoRA updates \citep{hu2022lora} in the hidden layers of the action model
(Figure~\ref{fig:policyweave-overview}(a)).
Existing merging methods can then combine these updates without also merging
task-specific action interfaces.

\begin{figure}[t]
  \centering
  \includegraphics[width=0.95\linewidth]{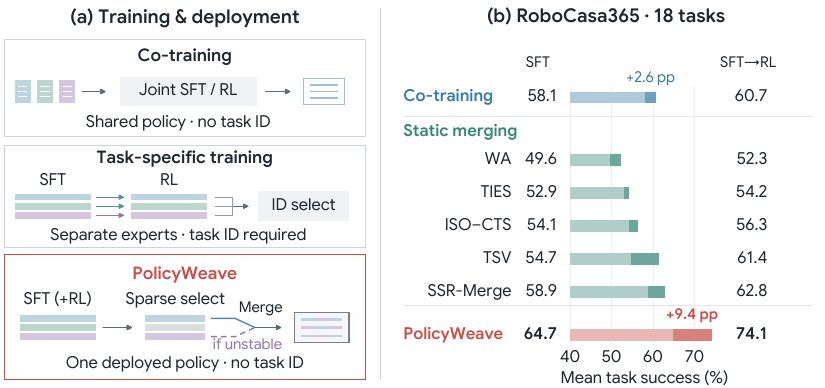}
  \caption{\textbf{(a)} Co-training, task-specific adaptation, and \policyweave. \textbf{(b)} Mean success (\%) on 18 RoboCasa365 tasks with SFT and SFT then RL, comparing co-training with model merging.}
  \label{fig:closest-method-comparison}
\end{figure}

A shared action interface improves mergeability, but does not eliminate
interference between task updates. Widely used static merging
methods combine these updates into one set of parameters for all tasks
\citep{ilharco2023taskarithmetic,yadav2023ties,gargiulo2025tsv}, so updates
unrelated to the current task can still affect its execution.
Selecting relevant updates based on the input offers a way to reduce
this interference
\citep{tang2024wemoe,lu2024twinmerging}.
In VLAs, observations and instructions are encoded into visual-language
context that conditions the action model, providing a basis for identifying
relevant experts. However, tasks sharing objects or goals can produce similar
contexts: selecting one expert risks choosing the wrong policy, while merging
several can introduce conflicting updates. Our insight is that the experts'
own responses to this context can guide expert selection. \policyweave uses
these responses to rank experts and ranking stability to decide whether to
select one expert or merge a sparse set, without a trained router
or calibration trajectories (Figure~\ref{fig:policyweave-overview}(b)).

We evaluate \policyweave with GR00T N1.5 \citep{bjorck2025groot}
on 18 atomic RoboCasa365 \citep{nasiriany2026robocasa365} tasks,
using only 10\% of the target-task demonstrations for SFT.
Preserving the shared action interface retains individual expert
performance (65.6\% versus 64.1\% with task-specific interfaces), while
raising the average success rate across four static merging methods from
17.0\% to 52.8\% (Figure~\ref{fig:action-interface-motivation}(a)).
Using the same shared-interface SFT experts, context-guided sparse merging
further improves success to 64.7\%. Task-specific RL then raises merged
success to 74.1\%, while joint RL improves the co-training baseline
from 58.1\% to 60.7\%
(Figure~\ref{fig:closest-method-comparison}(b)).
Experiments on 10 long-horizon LIBERO-10 \citep{liu2023libero} tasks show
retention of multi-stage control capabilities. On four Piper-arm manipulation
tasks, \policyweave also outperforms the co-trained baselines, supporting
multi-task deployment of independently learned skills on a physical robot. These experiments
support separating task-specific optimization from multi-task deployment:
experts can be trained independently and merged for deployment by preserving
their compatibility during post-training and controlling task interference
when merging.
\par\endgroup

\afterpage{\flushbottom\setlength{\textfloatsep}{\introSavedTextFloatSep}}
\begingroup
\setlength{\parskip}{3pt}
\section{Method: PolicyWeave}
\label{sec:method}

\begin{figure}[!t]
  \centering
  \includegraphics[width=\linewidth]{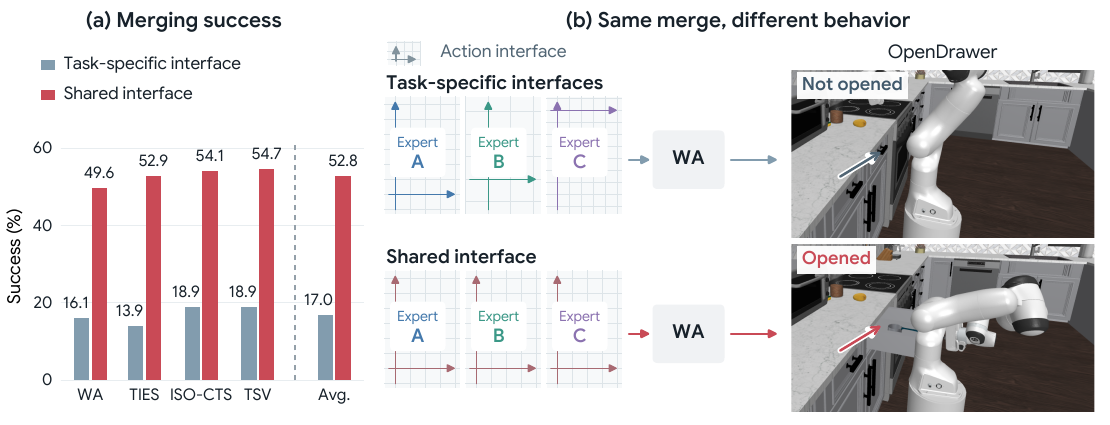}
  \caption{\textbf{A shared action interface supports effective merging of experts
  trained on 18 RoboCasa365 tasks.}
  (a) Mean SFT success across these tasks for four static merging methods;
  Avg. denotes their unweighted mean.
  (b) Different action interfaces, the same weight-averaging (WA) rule.
  Paired OpenDrawer
  rollouts show successful opening only with the shared interface.}
  \label{fig:action-interface-motivation}
\end{figure}

\afterpage{\afterpage{%
\begin{figure}[!t]
  \centering
  \includegraphics[width=\linewidth]{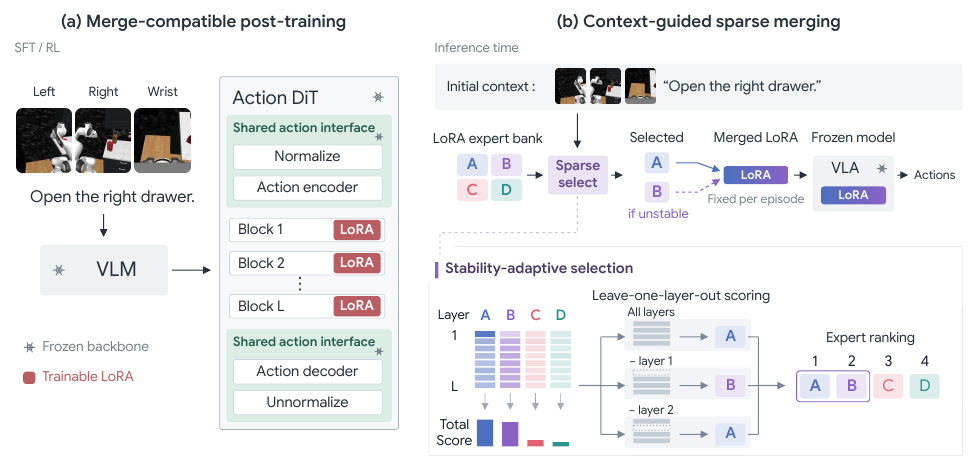}
  \caption{Left: \policyweave trains task-specific LoRA experts while freezing
  the VLM and shared action interface. Right: initial context guides sparse
  expert selection and merging. Leave-one-layer-out stability determines the
  expert set, and merged parameters remain fixed per episode.}
  \label{fig:policyweave-overview}
\end{figure}
}}

\subsection{Merge-Compatible VLA Post-Training}
\label{sec:policyweave}

We consider $K$ target tasks indexed by $\taskset=\{1,\ldots,K\}$, with one
LoRA expert independently post-trained for each task from a common base VLA
$\baseparams$. The final expert parameters are
$\taskparams{t}\in\{\taskparams{t}^{\mathrm{SFT}},\taskparams{t}^{\mathrm{SFT+RL}}\}$,
obtained through SFT \citep{kim2025openvla} or SFT then RL
\citep{chen2025pirl}. In our setting, RL continues training the same
task-specific LoRA parameters. In both cases,
$\taskdelta{t}=\taskparams{t}-\baseparams$ collects the induced LoRA weight
updates relative to the common base. A merging
method $\mathcal{M}$ combines these updates into a deployed policy with
parameters
\begin{equation}
  \mergedparams
  =\baseparams+\mathcal{M}(\{\taskdelta{t}\}_{t\in\taskset}).
  \label{eq:generic-merge}
\end{equation}
Existing LoRA post-training configurations differ in which parts of a VLA
they adapt. For example, LoRA can adapt both the VLM and action transformer
in $\pi_0$ \citep{black2025pi0} and $\pi_{0.5}$ \pageSafeCitep{black2025pi05},
whereas the N1.5 configuration of GR00T \citep{bjorck2025groot} adapts the
action head while keeping the VLM frozen.

For merging, task adaptation should also preserve a common mapping between
model representations and physical actions. We call the normalization
statistics and the action head's input-output projectors, namely its action
encoder and decoder, the \emph{action interface}.
Task-specific normalization changes how physical actions are represented
within the model and how its predictions are converted back into physical
commands. Updating the encoder and decoder further alters the mappings
between actions and hidden representations.
An expert can compensate for interface changes through its internal weights,
but merging independently adapted weights can break this compensation.
A simple example in Appendix~\ref{app:shared-interface-analysis} shows that
parameter averaging can change the output even when the original experts
produce identical actions.

\paragraph{Evidence from static merging.} Figure~\ref{fig:action-interface-motivation}(a) compares four static merging
methods on 18 RoboCasa365 tasks: their average success rate increases from
17.0\% with task-specific interfaces to 52.8\% with the shared interface.
Figure~\ref{fig:action-interface-motivation}(b) illustrates these interface
differences with paired OpenDrawer rollouts: under the same WA rule,
the shared-interface policy successfully opens the drawer, while the
task-specific-interface policy produces actions that deviate
from the intended motion.

\paragraph{Post-training with a shared interface.} Motivated by these results, we propose \emph{merge-compatible VLA post-training}
(Figure~\ref{fig:policyweave-overview},left). Within a shared embodiment and action
space, the broadly trained base provides action statistics and input-output
mappings that can be reused across downstream tasks. \policyweave retains
its normalization statistics and freezes its action encoder, action decoder,
and VLM. Task adaptation is restricted to LoRA parameters in the hidden
action-model layers.
This post-training protocol is independent of the merging method: the same expert bank supports
existing static methods and the context-guided sparse merging described
next.

\subsection{Context-Guided Sparse Merging}
\label{sec:contextvote}

Even with a shared action interface, static merging can introduce interference
from task-irrelevant updates \citep{yadav2023ties,tang2024wemoe}.
As shown in Figure~\ref{fig:policyweave-overview}(b), \policyweave uses the
initial observation and instruction to select and merge
relevant expert updates. We consider tasks covered by the expert bank,
without a supplied task identity.

\paragraph{Merging selected expert updates.}
Let $c$ be the visual-language context encoded from the initial observation
and instruction, $\mathcal{I}^\star(c)\subseteq\taskset$ the selected expert
set, and $\gamma_t(c)$ its nonnegative merging weights that sum to one.
We merge the selected updates at every adapted hidden action-model layer
$u\in\mathcal{H}$:
\begin{equation}
  \Delta \mathbf{W}_{u}(c)
  =\sum_{t\in\mathcal{I}^\star(c)}\gamma_t(c)\Delta \mathbf{W}_{t,u}.
  \label{eq:contextvote-compose}
\end{equation}
Here $\mathcal{H}$ denotes the adapted layers, $\Delta \mathbf{W}_{t,u}$ is
expert $t$'s update, and $\Delta \mathbf{W}_u(c)$ is the merged update at layer
$u$. We add these updates to the base weights, using the same merging weights
across layers. With one selected expert, its weight is one and the deployed
parameters equal that expert's parameters. The initial context determines the
merged parameters, which remain fixed throughout the episode while the policy
responds to new observations. We omit the
context argument $c$ from subsequent notation.

\paragraph{Expert relevance from context responses.}
The frozen VLM encodes the instruction and observed scene into a common
context $c$ for all experts. This context conditions the action DiT through
cross-attention \citep{bjorck2025groot}. Task-specific key updates change
how action queries match this context, so we use their responses to it as
relevance scores. One base-model forward pass with task adapters disabled
provides the shared key-projection inputs, avoiding a full forward pass
through each expert.
Let $\mathbf{H}_\ell\in\mathbb{R}^{d_{\mathrm{in}}\times N}$ contain the
$N$ valid context-token representations as columns at the input to layer
$\ell$'s key projection, and let
$\Delta\mathbf{W}^{\mathrm{K}}_{t,\ell}\in\mathbb{R}^{d_{\mathrm{out}}\times d_{\mathrm{in}}}$
be expert $t$'s LoRA update to that projection. Here $d_{\mathrm{in}}$ and
$d_{\mathrm{out}}$ are the projection's input and output dimensions.
We normalize its response by both update and context magnitude
to compare experts independently of scale. For the set of scoring layers $\mathcal{C}$,
the layer score $e_{t,\ell}$ and aggregate expert score $s_t$ are scalars:
\begin{equation}
  e_{t,\ell}
  =
  \frac{
    \lVert\Delta\mathbf{W}^{\mathrm{K}}_{t,\ell}\mathbf{H}_\ell\rVert_F^2
  }{
    \lVert\Delta\mathbf{W}^{\mathrm{K}}_{t,\ell}\rVert_F^2
    \lVert\mathbf{H}_\ell\rVert_F^2
  },
  \qquad
  s_t=\sum_{\ell\in\mathcal{C}}e_{t,\ell}.
  \label{eq:contextvote-score}
\end{equation}
The score measures alignment between the context and the input directions
to which the key update is sensitive. All scoring layers contribute equally.

\paragraph{Expert selection by ranking stability.}
Selecting one expert risks choosing the wrong policy, while selecting several
can introduce conflicting updates. We determine the set size by checking
whether the highest-scoring expert remains unchanged when individual scoring
layers are excluded.
Let $\rho=(\rho_1,\ldots,\rho_K)$ be the expert indices sorted by descending
aggregate score, so $s_{\rho_1}\geq\cdots\geq s_{\rho_K}$.
We remove each layer's score contribution in turn and identify the expert
with the highest remaining score, without additional forward passes.
For layer $\ell$, this gives
\begin{equation}
  s_t^{(-\ell)}=s_t-e_{t,\ell},
  \qquad
  t_\ell^\star=\arg\max_{t\in\taskset}s_t^{(-\ell)}.
  \label{eq:contextvote-leave-one-out}
\end{equation}
Here $t_\ell^\star$ identifies the highest-scoring expert after removing
layer $\ell$. If $\rho_1$ remains first after every removal, we select it
alone. Otherwise, we collect all experts that become first under a layer
removal. We select the smallest top-$k$ set in the original ranking that
includes all these experts:
\begin{equation}
  k^\star
  =\max_{\ell\in\mathcal{C}}\operatorname{rank}_{\rho}(t_\ell^\star),
  \qquad
  \mathcal{I}^\star=\{\rho_1,\ldots,\rho_{k^\star}\},
  \label{eq:contextvote-adaptive-set}
\end{equation}
where $\operatorname{rank}_{\rho}(t_\ell^\star)$ is its position in the
original ranking, starting at 1. Fixed Top-$k$ variants use a prescribed
set size with the same ranking and weighting rule. In both cases, the
selected updates are weighted using the original aggregate scores:
$\gamma_t=s_t/\sum_{j\in\mathcal{I}^\star}s_j$ for $t\in\mathcal{I}^\star$.
Appendix~\ref{app:selection-stability} provides a theoretical justification
for the ranking-stability test.
Inference details and scorer ablations are given in
Appendices~\ref{app:contextvote-inference}
and~\ref{app:contextvote-scorer-ablation}.

\section{Experiments}
\label{sec:experiments}
\setlength{\textfloatsep}{10pt plus 2pt minus 2pt}
\setlength{\floatsep}{10pt plus 2pt minus 2pt}
We organize our evaluation around three questions: (1) Can merging preserve
independently acquired task capabilities, including gains from task-specific
RL and long-horizon control? (2) How does a shared action interface improve
merging? (3) Can context responses guide which experts to select and how many
to merge?

\subsection{Experimental Setup}
\label{sec:experimental-setup}
\paragraph{Baselines and evaluation.}
We use two simulation benchmarks to examine complementary aspects of VLA merging.
On 18 atomic RoboCasa365 tasks \citep{nasiriany2026robocasa365}, SFT and
SFT then RL expert banks test both skill retention and whether improvements
from task-specific RL survive multi-task deployment. LIBERO-10
\citep{liu2023libero} instead tests retention of long-horizon control using
an SFT-only bank, with each expert trained on a complete multi-stage task.
We compare \policyweave with five representative post-hoc merging baselines:
weight averaging (WA) \citep{wortsman2022modelsoups}, TIES
\mbox{\citep{yadav2023ties}}, ISO-CTS \citep{marczak2025isotropic}, TSV-Merge
\citep{gargiulo2025tsv}, and SSR-Merge \citep{wei2026ssrmerge}. Together, they
cover direct averaging, coordinate-level conflict resolution, and
subspace-based merging, including methods tailored to LoRA.
We evaluate shared-interface post-training across four static merging methods
(Figure~\ref{fig:action-interface-motivation}). Our main merging comparisons
then hold the shared-interface expert bank fixed within each benchmark.
For each task, we report the mean success rate over 50
evaluation episodes. For the challenging long-horizon LIBERO-10 tasks, we
additionally report partial success, measured by the maximum fraction of goal
predicates simultaneously satisfied during each episode and averaged across
episodes.
\paragraph{Implementation details.}
Our expert banks use GR00T N1.5 \citep{bjorck2025groot}, initialized from a
benchmark-specific base checkpoint. We train rank-8 LoRA experts in the action
model while preserving the \policyweave action interface. Each RoboCasa365
SFT expert uses 50 demonstrations, or
10\% of the available data, and RL follows the Flow-SDE formulation of
$\pi_{\mathrm{RL}}$ \citep{chen2025pirl}. The RoboCasa365 \textit{Co-training}
baseline starts from the official checkpoint co-trained on all 18 tasks using
10\% of the training data. We continue this policy with 600 RL steps,
comparable to the 610 steps summed across the task-specific continuations.
LIBERO-10 uses an SFT expert bank with 38 demonstrations per task on average.
\policyweave uses eight action
cross-attention key projections for scoring and fixes the merged update for the
episode. Complete protocols are given in
Appendix~\ref{app:experimental-details}.

\begin{table}[!t]
\setlength{\abovecaptionskip}{3pt}
\setlength{\belowcaptionskip}{0pt}
\caption{\textbf{RoboCasa365 success (\%), averaged over 50 episodes per task.}
Each pair reports \mbox{\textbf{SFT $\rightarrow$ SFT then RL}}. All merging methods use
the same shared-interface expert bank. Bold marks the best merging result
for each task and stage.}
\label{tab:robocasa-main}
\centering
\small
\setlength{\tabcolsep}{1.7pt}
\renewcommand{\arraystretch}{1.03}
\definecolor{robogain}{HTML}{8F3039}
\newcommand{\roboheader}[1]{\makebox[\linewidth][c]{#1}}
\newcommand{\roboarrow}{\scalebox{0.65}[1]{$\rightarrow$}}
\newcommand{\robopair}[2]{\makebox[\linewidth][c]{%
  \makebox[1.05em][r]{#1}\kern0.3pt\roboarrow\kern0.3pt\makebox[1.05em][l]{#2}}}
\newcommand{\robomean}[1]{\makebox[\linewidth][c]{#1}}
\newcommand{\robogain}[1]{\robomean{\textcolor{robogain}{#1}}}
\begin{tabularx}{\textwidth}{@{}p{0.22\textwidth}
  >{\columncolor{gray!15}\centering\arraybackslash}X
  >{\centering\arraybackslash}X
  !{\hspace{2pt}}*{5}{>{\centering\arraybackslash}X}
  >{\columncolor{policyweaveblue}\centering\arraybackslash}X@{}}
\toprule
& \multicolumn{2}{c}{Baselines}
& \multicolumn{6}{c}{Merging methods} \\
\cmidrule(lr){2-3}\cmidrule(lr){4-9}
Task & \cellcolor{white}\roboheader{\textit{Indiv.}} &
\roboheader{\shortstack{\textit{Co-}\\\textit{training}}} &
\roboheader{WA} & \roboheader{TIES} &
\roboheader{\shortstack{ISO-\\CTS}} &
\roboheader{TSV} & \roboheader{\shortstack{SSR-\\Merge}} &
\cellcolor{white}\roboheader{\shortstack{\textbf{Policy}\\\textbf{Weave}}} \\
\midrule
Close blender lid & \robopair{22}{32} & \robopair{8}{24} & \robopair{8}{8} & \robopair{8}{18} & \robopair{10}{0} & \robopair{4}{10} & \robopair{14}{10} & \robopair{\textbf{20}}{\textbf{30}} \\
Close fridge & \robopair{68}{74} & \robopair{64}{70} & \robopair{74}{74} & \robopair{66}{78} & \robopair{62}{68} & \robopair{\textbf{78}}{68} & \robopair{70}{\textbf{82}} & \robopair{\textbf{78}}{76} \\
Close toaster oven & \robopair{80}{80} & \robopair{68}{76} & \robopair{46}{56} & \robopair{54}{40} & \robopair{68}{\textbf{76}} & \robopair{68}{68} & \robopair{46}{44} & \robopair{\textbf{78}}{\textbf{76}} \\
Set up coffee mug & \robopair{50}{76} & \robopair{50}{40} & \robopair{26}{26} & \robopair{28}{32} & \robopair{40}{54} & \robopair{32}{50} & \robopair{30}{52} & \robopair{\textbf{56}}{\textbf{70}} \\
Navigate kitchen & \robopair{22}{68} & \robopair{28}{28} & \robopair{2}{6} & \robopair{6}{10} & \robopair{8}{14} & \robopair{14}{22} & \robopair{6}{6} & \robopair{\textbf{20}}{\textbf{62}} \\
Open cabinet & \robopair{86}{94} & \robopair{72}{86} & \robopair{60}{58} & \robopair{60}{66} & \robopair{66}{80} & \robopair{68}{64} & \robopair{74}{78} & \robopair{\textbf{80}}{\textbf{88}} \\
Open drawer & \robopair{70}{84} & \robopair{80}{80} & \robopair{60}{66} & \robopair{70}{70} & \robopair{66}{76} & \robopair{78}{72} & \robopair{68}{74} & \robopair{\textbf{82}}{\textbf{82}} \\
Open mixer head & \robopair{94}{96} & \robopair{82}{90} & \robopair{76}{76} & \robopair{74}{74} & \robopair{74}{68} & \robopair{68}{80} & \robopair{88}{84} & \robopair{\textbf{92}}{\textbf{92}} \\
Counter $\to$ cabinet & \robopair{64}{74} & \robopair{64}{62} & \robopair{54}{68} & \robopair{44}{58} & \robopair{40}{60} & \robopair{62}{64} & \robopair{\textbf{74}}{64} & \robopair{56}{\textbf{70}} \\
Counter $\to$ stove & \robopair{70}{70} & \robopair{60}{48} & \robopair{56}{70} & \robopair{68}{70} & \robopair{64}{56} & \robopair{\textbf{72}}{\textbf{78}} & \robopair{\textbf{72}}{76} & \robopair{60}{70} \\
Drawer $\to$ counter & \robopair{36}{58} & \robopair{44}{40} & \robopair{30}{18} & \robopair{\textbf{34}}{36} & \robopair{16}{24} & \robopair{28}{36} & \robopair{26}{40} & \robopair{\textbf{34}}{\textbf{60}} \\
Sink $\to$ counter & \robopair{76}{76} & \robopair{74}{74} & \robopair{68}{78} & \robopair{76}{78} & \robopair{68}{70} & \robopair{84}{82} & \robopair{\textbf{90}}{\textbf{88}} & \robopair{72}{72} \\
Toaster $\to$ counter & \robopair{84}{84} & \robopair{66}{64} & \robopair{\textbf{82}}{\textbf{90}} & \robopair{70}{66} & \robopair{78}{82} & \robopair{68}{78} & \robopair{74}{88} & \robopair{78}{82} \\
Slide dishwasher rack & \robopair{86}{86} & \robopair{58}{66} & \robopair{66}{54} & \robopair{66}{52} & \robopair{76}{70} & \robopair{54}{62} & \robopair{72}{74} & \robopair{\textbf{78}}{\textbf{84}} \\
Turn off stove & \robopair{50}{60} & \robopair{22}{30} & \robopair{14}{22} & \robopair{44}{22} & \robopair{\textbf{46}}{38} & \robopair{26}{50} & \robopair{42}{40} & \robopair{42}{\textbf{56}} \\
Turn on kettle & \robopair{74}{90} & \robopair{70}{70} & \robopair{\textbf{96}}{\textbf{98}} & \robopair{88}{94} & \robopair{92}{90} & \robopair{88}{96} & \robopair{90}{84} & \robopair{90}{86} \\
Turn on microwave & \robopair{70}{82} & \robopair{56}{64} & \robopair{40}{42} & \robopair{44}{60} & \robopair{38}{32} & \robopair{44}{60} & \robopair{50}{60} & \robopair{\textbf{64}}{\textbf{90}} \\
Turn on sink faucet & \robopair{78}{88} & \robopair{80}{80} & \robopair{34}{32} & \robopair{52}{52} & \robopair{62}{56} & \robopair{48}{66} & \robopair{74}{86} & \robopair{\textbf{84}}{\textbf{88}} \\
\midrule
\textbf{Mean SFT} & \robomean{65.6} & \robomean{58.1} & \robomean{49.6} & \robomean{52.9} & \robomean{54.1} & \robomean{54.7} & \robomean{58.9} & \robomean{\textbf{64.7}} \\
\textbf{Mean SFT then RL} & \robomean{76.2} & \robomean{60.7} & \robomean{52.3} & \robomean{54.2} & \robomean{56.3} & \robomean{61.4} & \robomean{62.8} & \robomean{\textbf{74.1}} \\
Gain (pp) & \robogain{+10.6} & \robogain{+2.6} & \robogain{+2.7} & \robogain{+1.3} & \robogain{+2.2} & \robogain{+6.7} & \robogain{+3.9} & \robogain{+9.4} \\
\bottomrule
\end{tabularx}
\end{table}

\Needspace{5\baselineskip}
\subsection{Preserving Task Capabilities through Merging}
\paragraph{Atomic Skills and RL Gains (RoboCasa365).}
Table~\ref{tab:robocasa-main} compares \policyweave with static merging methods
on the same SFT and SFT then RL expert banks. SSR-Merge is the strongest static
method in both banks, reaching 58.9\% and 62.8\% success, respectively.
\policyweave reaches 64.7\% and 74.1\%, closer to the corresponding
\textit{Individual} results of 65.6\% and 76.2\%.
\par
\Needspace{14\baselineskip}
\suppressfloats[t]
\begingroup
\setlength{\columnsep}{12pt}
\setlength{\intextsep}{5pt}
\begin{wrapfigure}{r}{0.46\textwidth}
  \centering
  \setlength{\abovecaptionskip}{3pt}
  \setlength{\belowcaptionskip}{0pt}
  \includegraphics[width=\linewidth]{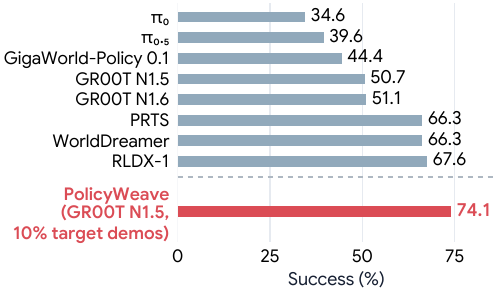}
  \caption{\protect\raggedright RoboCasa365
  Atomic-Seen leaderboard references.}
  \label{fig:robocasa-leaderboard}
\end{wrapfigure}
The change from SFT to SFT then RL tests how much of the experts' improvement
survives merging. Task-specific RL raises \textit{Individual} success by
10.6\%. All static methods benefit from the improved experts, but retain
only part of this gain. \policyweave retains a larger gain of 9.4\%, whereas
joint RL improves \textit{Co-training} by 2.6\% with a comparable RL-step
budget. Independent RL can therefore improve multi-task deployment, provided
the merging method preserves the resulting capabilities.
For reference, Figure~\ref{fig:robocasa-leaderboard} also includes selected
RoboCasa365 \citep{nasiriany2026robocasa365} Atomic-Seen leaderboard results
reported under their distinct protocols.
\par
\WFclear
\endgroup

\begin{figure}[t]\vspace{-3ex}
  \centering
  \setlength{\abovecaptionskip}{3pt}
  \setlength{\belowcaptionskip}{0pt}
  \includegraphics[width=\textwidth]{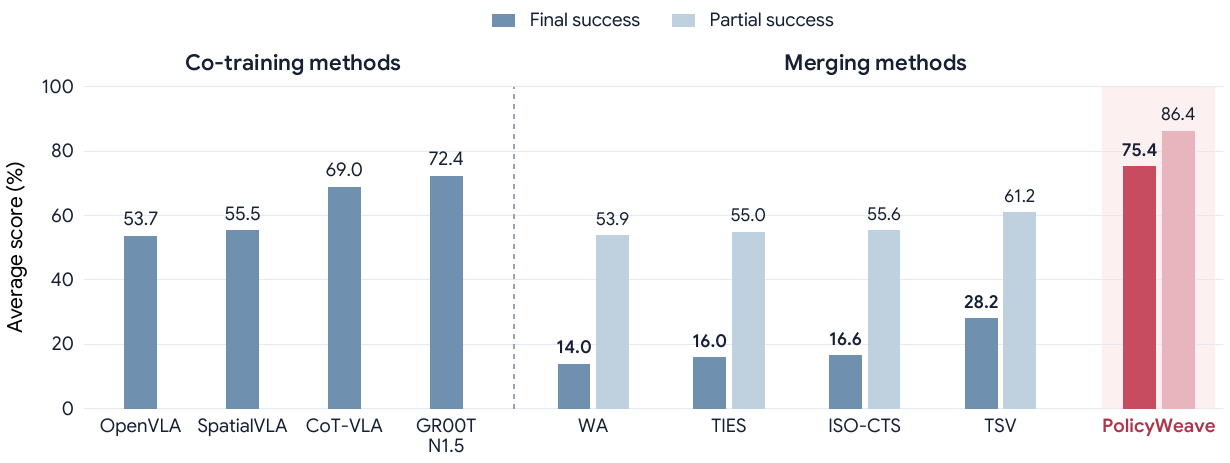}\vspace{-2ex}
  \caption{\textbf{Long-horizon performance on LIBERO-10.}
Left: published results for OpenVLA \citep{kim2025openvla},
SpatialVLA \citep{qu2025spatialvla}, CoT-VLA \citep{zhao2025cotvla},
and co-trained GR00T N1.5 baseline.
Right: success rates across merging methods
using the same expert bank.}
  \label{fig:libero-main}
\end{figure}

\Needspace{5\baselineskip}
\begingroup
\setlength{\columnsep}{10pt}
\setlength{\intextsep}{3pt}
\setlength{\parskip}{2pt}
\begin{wraptable}[6]{r}{0.43\linewidth}
  \vspace{-4pt}
  \raggedleft
  \setlength{\abovecaptionskip}{3pt}
  \setlength{\belowcaptionskip}{3pt}
  \caption{LIBERO-10 merging retention.}
  \label{tab:libero-merging-retention}
  \setlength{\tabcolsep}{3pt}
  \renewcommand{\arraystretch}{1.03}
  \begin{tabular}{@{}lcc@{}}
    \toprule
    Method & Indiv. $\rightarrow$ Merged & Ret. $\uparrow$ \\
    \midrule
    \textsc{MergeVLA} & $89.6 \rightarrow 70.8$ & 79.0 \\
    \rowcolor{policyweaveblue} \policyweave & $80.8 \rightarrow 75.4$ & \textbf{93.3} \\
    \bottomrule
  \end{tabular}
\end{wraptable}

\paragraph{Long-Horizon Skill Retention (LIBERO-10).}
\begin{figure}[t]
  \setlength{\abovecaptionskip}{3pt}
  \setlength{\belowcaptionskip}{0pt}
  \centering
  \includegraphics[width=\textwidth]{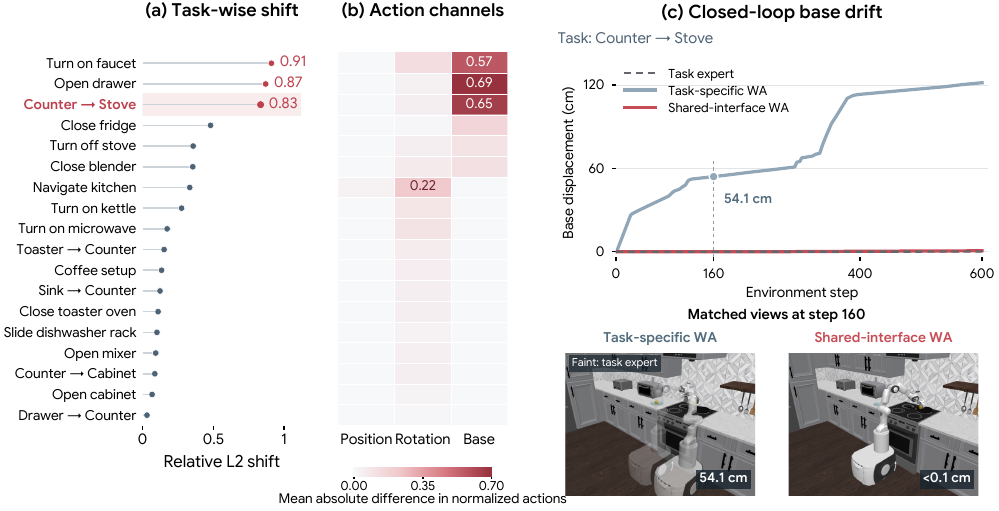}
  \caption{\textbf{Action normalization and closed-loop base drift on RoboCasa365.}
  (a) Differences between task-specific and shared normalization of the same
  raw actions. (b) The largest differences occur in base-motion channels;
  rows follow (a). (c) Actual base displacement from a matched initial scene
  on PickPlaceCounterToStove. The views show step 160.}
  \label{fig:p92-coordinate-geometry}
\end{figure}
LIBERO-10 tests whether merging preserves complete long-horizon behaviors.
Without a shared action interface, all four static methods yield 0\% final
success (Appendix~\ref{app:mergeability-metrics}); with it, they reach
14.0--28.2\%. Figure~\ref{fig:libero-main} also reports partial success,
which reaches 53.9--61.2\% for these methods. The shared interface improves
mergeability, but much of the intermediate progress still fails to translate
into full task completion, consistent with accumulated control errors over
longer sequences.
\policyweave achieves 75.4\% final success and 86.4\% partial success,
largely preserving both intermediate progress and long-horizon task completion.
Its final success also exceeds the selected cotrained VLA results, including GR00T N1.5 at 72.4\%.
\par
To further assess skill retention, we compare \policyweave with our MergeVLA
reproduction on LIBERO-10. Since their VLA backbones and post-training pipelines
differ, we report merged \mbox{success} as a fraction of each method's
\textit{Individual} expert success. \policyweave retains 93.3\%, compared with
79.0\% for MergeVLA (Table~\ref{tab:libero-merging-retention}).
\par
\WFclear
\endgroup
\subsection{The Role of a Shared Action Interface}
\label{sec:action-coordinate-analysis}
\begingroup
\setlength{\columnsep}{10pt}
\setlength{\intextsep}{3pt}
The merging results on RoboCasa365 (Figure~\ref{fig:action-interface-motivation})
and LIBERO-10 motivate examining how action-interface differences affect
physical execution. We first apply task-specific and shared normalization to
the same 1,800 raw actions from 18 RoboCasa365 tasks.
Figure~\ref{fig:p92-coordinate-geometry}(a,b) shows that differences in normalized
actions vary across tasks and are largest in the base-motion
channels. Guided by this observation, we examine base motion during closed-loop execution on
PickPlaceCounterToStove, comparing a task expert with WA policies constructed
from the task-specific-interface and shared-interface banks. The policies
start from matched scenes with paired denoising noise. 
\begin{wraptable}[10]{r}{0.4\textwidth}
  \centering
  \setlength{\abovecaptionskip}{2pt}
  \setlength{\belowcaptionskip}{2pt}
  \caption{Sensitivity of individual experts to action-interface changes.}\vspace{1ex}
  \label{tab:interface-component-interventions}
  \small
  \setlength{\tabcolsep}{2pt}
  \renewcommand{\arraystretch}{1.10}
  \begin{tabularx}{\linewidth}{@{}Xr@{}}
    \toprule
    Setting & Success (\%) \\
    \midrule
    \rowcolor{gray!12}Individual & 64.1 \\
    Encoder WA & 17.1~(\textcolor{darkred}{$-47.0\%$}) \\
    Decoder WA & 60.4~(\textcolor{darkred}{$-3.7\%$}) \\
    Base Normalization & 12.3~(\textcolor{darkred}{$-51.8\%$}) \\
    \bottomrule
  \end{tabularx}
\end{wraptable}
 In the scene shown in
Figure~\ref{fig:p92-coordinate-geometry}(c), task-specific WA produces
substantial base drift, reaching 54.1\,cm by step 160, whereas shared-interface
WA stays within 0.1\,cm of its initial position. 

To examine how individual interface components affect control, we use experts
trained with task-specific action interfaces and replace
each expert's action encoder or decoder with its average across
experts, or replace its action normalization with the base statistics,
keeping all other components unchanged. Encoder averaging and normalization
replacement reduce success more than decoder averaging
(Table~\ref{tab:interface-component-interventions}). While these components
differ in sensitivity, keeping the full interface shared ensures that merged
hidden-layer updates use the same action input and output mappings.
\par\WFclear
\endgroup

\begin{table}[t]
  \vspace{-12pt}
  \centering
  \caption{\textbf{LIBERO-10 selection and merging (500 episodes).}
  Accuracy and success are in \%; $\bar{k}$ is the mean number of selected experts.
  The three shared-interface \policyweave variants use the same context-response scorer.
  Subset success uses cases where adaptive selection merges multiple experts,
  defined separately for each expert bank.
  }
  \label{tab:libero-selection-main}

  \small
  \setlength{\tabcolsep}{4pt}
  \renewcommand{\arraystretch}{1.08}
  \newcommand{\pwselection}[1]{\policyweave (#1)}

  \begin{tabular}{@{}lrrrr@{}}
    \toprule
    Method & Top-1 Acc. & Mean $k$ & Success (all) & Subset success \\
    \midrule
    Trainable MLP
      & 89.4 & 1 & 69.4 & -- \\
    \pwselection{Top-1}
      & 93.6 & 1 & 73.2 & 36.4 \\
    \pwselection{Top-2}
      & 93.6 & 2 & 66.2 & 57.6 \\
    \rowcolor{policyweaveblue}
    \policyweave
      & 93.6 & 1.076 & \textbf{75.4} & \textbf{63.6} \\
    \addlinespace[3pt]
    \policyweave w/o shared interface
      & 90.2 & 1.112 & 70.0 & 0.0 \\
    \bottomrule
  \end{tabular}
\end{table}

\begin{figure}[!t]
  \centering
  \setlength{\abovecaptionskip}{3pt}
  \setlength{\belowcaptionskip}{0pt}
  \begin{minipage}[c]{0.43\linewidth}
    \centering
    \includegraphics[height=72pt]{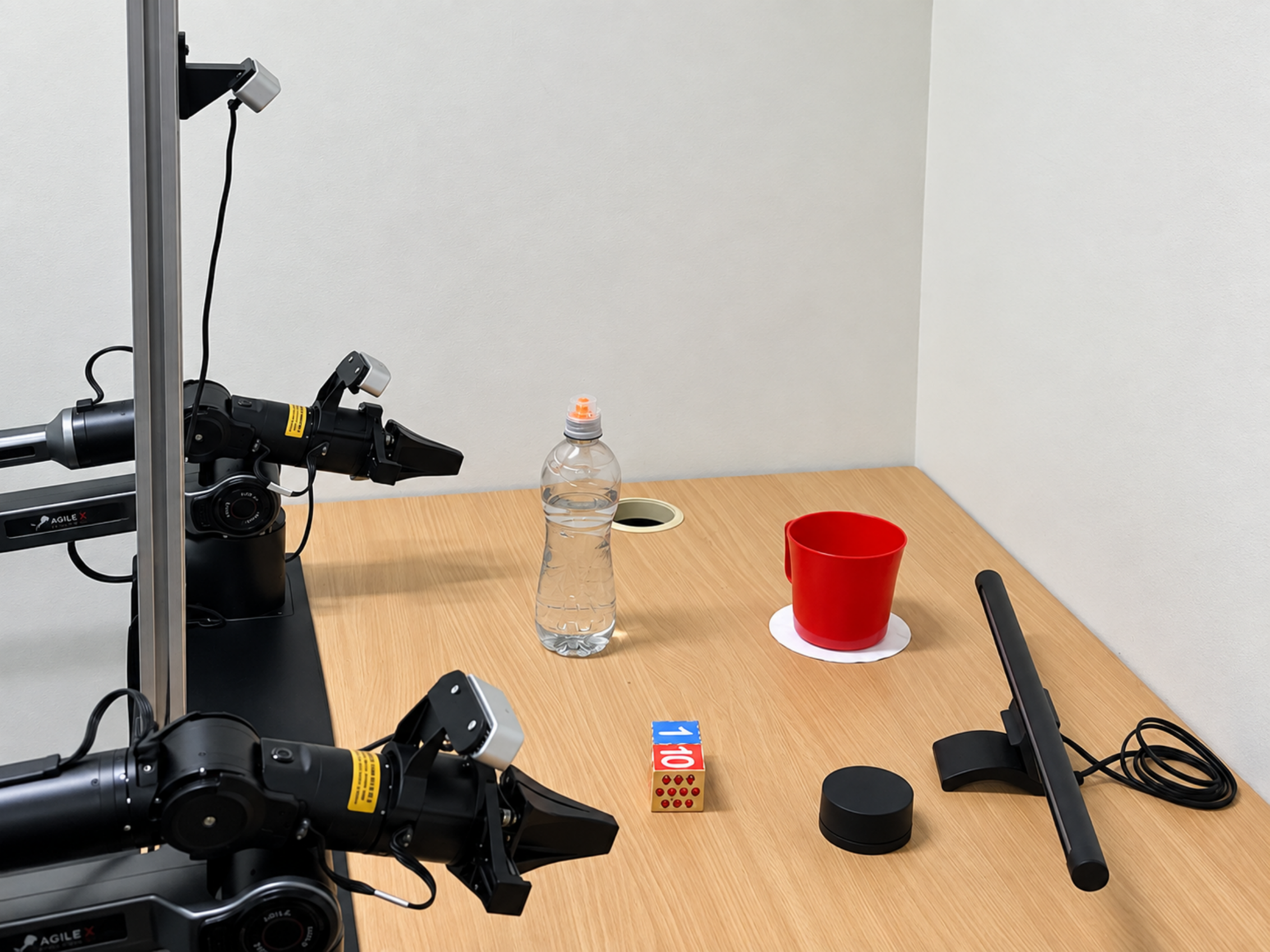}
    \caption{\textbf{Real-world evaluation setup.}
Experiments use an AgileX Piper arm and two cameras across 4 manipulation tasks.}
    \label{fig:real-robot-setup}
  \end{minipage}\hfill
  \begin{minipage}[c]{0.55\linewidth}
\makeatletter
\def\@captype{table}
\makeatother
  \centering
  \setlength{\abovecaptionskip}{0pt}
  \caption{\textbf{Real-robot success (\%) over 20 rollouts per task and method.}
  Mean is the four-task average.}
  \label{tab:real-robot-results}
  \small
  \setlength{\tabcolsep}{2pt}
  \renewcommand{\arraystretch}{1.08}
  \begin{tabular*}{\linewidth}{@{\extracolsep{\fill}}lccc@{}}
    \toprule
    Task & $\pi_{0.5}$ & GR00T N1.5 & \policyweave \\
    \midrule
    Upright bottle & 30 & 10 & \cellcolor{policyweaveblue}\textbf{40} \\
    Stack blocks & 15 & 10 & \cellcolor{policyweaveblue}\textbf{55} \\
    Press button & 80 & 85 & \cellcolor{policyweaveblue}\textbf{95} \\
    Place cup & 70 & 65 & \cellcolor{policyweaveblue}\textbf{80} \\
    \midrule
    Mean & 48.8 & 42.5 & \cellcolor{policyweaveblue}\textbf{67.5} \\
    \bottomrule
  \end{tabular*}

  \end{minipage}
  \par
\end{figure}

\subsection{Expert Selection and Sparse Merging}
\label{sec:contextvote-analysis}
\begingroup
\setlength{\columnsep}{12pt}
\setlength{\intextsep}{6pt}

\paragraph{Expert identification.}
We focus on LIBERO-10, where long-horizon tasks share objects and intermediate
goals, to test expert scoring and the choice of which updates to merge.
Table~\ref{tab:libero-selection-main} compares three shared-interface \policyweave variants
with a trainable MLP selector on the same expert bank. The MLP learns a context-to-expert
mapping, analogous to the learned routing in Twin-Merging
\citep{lu2024twinmerging}. Context-response scoring reaches
93.6\% Top-1 accuracy versus 89.4\% for the MLP. With single-expert selection
in both cases, \policyweave (Top-1) also achieves higher closed-loop success
than the MLP. The experts' own responses thus provide effective
selection signals without training a separate selection model.

\paragraph{Sparse merging.}
The shared-interface \policyweave variants share expert scores and differ only in the
selection rule. \policyweave (Top-2) lowers success relative to \policyweave (Top-1),
showing that including another expert can introduce harmful interference.
Stability-adaptive selection instead reaches 75.4\% success versus 73.2\%
for Top-1, with only 1.076 experts selected on average.
On the same episodes where adaptive selection uses $k>1$, it achieves
63.6\% success, compared with only 36.4\% for Top-1. These results
support adapting the number of merged updates to ranking stability rather
than always selecting one or two experts.
Applying the same selection and merging rule to experts trained without a
shared action interface yields 90.2\% Top-1 accuracy and 70.0\% overall
success. Expert identification remains effective, yet no rollout succeeds
in that variant's own multi-expert subset. Identifying relevant experts
therefore does not by itself ensure that their updates can be merged
successfully. These results support preserving the shared interface not
only for static merging, but also when sparse selection calls for merging
multiple experts. Further ablations are provided in
Appendix~\ref{app:additional-results}.
\par\endgroup
\Needspace{5\baselineskip}
\subsection{Real Robot Deployment}
\label{sec:real-robot}
We test independent post-training and merging on an AgileX Piper arm with
six revolute joints and a parallel gripper
(Figure~\ref{fig:real-robot-setup}). The four tasks are placing a bottle
upright, stacking two blocks, pressing a button to turn on a light, and
placing a cup on a coaster. We compare \policyweave with $\pi_{0.5}$
\citep{black2025pi05} and GR00T N1.5 \citep{bjorck2025groot} co-trained on
all four tasks, using 20 rollouts per task and method. Table~\ref{tab:real-robot-results} shows that \policyweave outperforms GR00T
co-training on all four tasks, with the largest gains on placing the bottle
upright and stacking blocks. These tasks require controlled object
reorientation or precise placement and are the most difficult for the
co-trained policy. With the same shared base and action interface,
\policyweave allows each expert to learn its task independently.
It achieves 100\% Top-1 expert-selection accuracy across the four tasks,
identifying the corresponding expert without a supplied task identity.
Together, these results support independent adaptation followed by
context-guided sparse merging as a way to deploy specialized skills that are harder
to acquire through co-training. Data and training settings are given in
Appendix~\ref{app:real-robot}.

\section{Related Work}
\label{sec:related-work}
\subsection{Multi-Task Vision-Language-Action Models}
\textbf{Pretraining and post-training.} Generalist vision-language-action (VLAs) \citep{brohan2023rt2,ghosh2024octo,kim2025openvla,
black2025pi0,black2025pi05,bjorck2025groot,shukor2025smolvla} combine heterogeneous robot data with vision-language
pretraining to support diverse tasks and embodiments.
Supervised fine-tuning (SFT) adapts these policies to downstream tasks using
expert demonstrations \citep{kim2025openvla,black2025pi0}.
RL further improves task performance through interaction
\citep{li2026simplevlarl,chen2025pirl,lyu2026fpo,wang2026qvgm,cao2026z1}.

\noindent\textbf{Structured multi-task skill learning.}
AtomicVLA \citep{zhang2026atomicvla} uses skill-guided action experts.
DyGRO-VLA \citep{lin2026dygrovla} routes grouped RL residuals across tasks.
DiTEA \citep{li2026ditea} uses language instructions to select experts within
the action model. KinRT \citep{yang2026kinrt} uses kinematic trajectory clusters
to supervise routing in an MoE action model.
These methods modify the training architecture, whereas \policyweave merges
independently post-trained updates without architectural changes.
\subsection{Model Merging}
Model merging combines task-adapted models without joint retraining.
Static methods produce one input-independent parameterization
\citep{wortsman2022modelsoups,matena2022fisher,ilharco2023taskarithmetic}.
They address conflicts through sign resolution, sparsification, and masking
\citep{yadav2023ties,yu2024dare,davari2024breadcrumbs,wang2024consensus}, or
subspace decomposition and projection
\citep{sun2025cat,gargiulo2025tsv,marczak2025isotropic}. Methods tailored to
LoRA search over adapter combinations or align low-rank updates
\citep{huang2024lorahub,stoica2025knots,zhao2025loralego,
panariello2025corespace,wei2026ssrmerge}.
Dynamic methods condition expert selection or merging on
the input
\citep{tang2024wemoe,lu2024twinmerging,oh2025dawin,wang2024loraflow}.
Policy perturbations can compound during closed-loop control, making VLA
merging particularly challenging
\citep{ross2011dagger,fu2026mergevla}.
CORAL \citep{luo2026coral} independently trains LoRA experts and folds one
into the base policy using a client-side instruction-to-expert function.
MergeVLA \citep{fu2026mergevla} redesigns the action expert and requires
task-specific masks even with static merging methods. \policyweave instead
preserves a shared action interface during post-training, making experts
compatible with existing static merging methods.
An extended comparison appears in Appendix~\ref{app:extended-related-work}.


\section{Conclusion}
\label{sec:conclusion}
Independent post-training allows task-specific optimization, but deploying the
resulting capabilities together requires compatible experts. \policyweave
preserves the action interface during post-training, improving compatibility
with existing static merging methods while retaining individual task performance.
Context-guided sparse merging limits the remaining interference at deployment.
RoboCasa365 experiments show that gains from independent RL carry into the
merged policy, while LIBERO-10 and Piper arm experiments support long-horizon
skill retention and real-world multi-task manipulation.
Together, these results show how maintaining compatibility during post-training
and controlling interference at deployment enable independently learned skills
to support a multi-task policy. Separating expert post-training from merging
also provides a basis for developing model merging methods for VLAs without
redesigning the policy.

\par\endgroup
\begingroup
\interlinepenalty=10000
\bibliography{refs}
\bibliographystyle{iclr2027_conference}
\endgroup
\clearpage
\appendix
\makeatletter
\setlength{\@fptop}{0pt}
\makeatother
\section*{Appendix Overview}
The appendix is organized as follows:
\begin{itemize}
  \setlength{\itemsep}{1pt}
  \setlength{\parskip}{0pt}
  \item Appendix~\ref{app:extended-related-work}: extended related work and
  comparisons with CORAL and MergeVLA.
  \item Appendix~\ref{app:contextvote-details}: action-interface analysis,
  context-response scoring, ranking stability, and the inference algorithm.
  \item Appendix~\ref{app:experimental-details}: checkpoints, training,
  evaluation protocols, and baseline configurations.
  \item Appendix~\ref{app:additional-results}: interface diagnostics,
  task-level results, expert selection, robustness, and scaling.
  \item Appendix~\ref{app:real-robot}: the Piper setup, training, and
  successful rollout sequences.
  \item Appendix~\ref{sec:limitations}: limitations and directions for future work.
\end{itemize}

\section{Extended Related Work}
\label{app:extended-related-work}

\subsection{Multi-Task and Modular VLA Learning}

Generalist VLAs acquire broad robot capabilities by sharing parameters across
heterogeneous datasets during pretraining or joint finetuning. This line spans
autoregressive action tokenization, continuous action generation, open
cross-embodiment datasets, and resource-efficient policies
\citep{brohan2023rt2,openx2023rtx,ghosh2024octo,kim2025openvla,
black2025pi0,black2025pi05,bjorck2025groot,shukor2025smolvla}.
Although scale and data diversity improve transfer, adding a new deployment
task generally entails updating a shared policy or repeating multi-task
training with the enlarged task set.

Modular approaches introduce explicit structure for specialization. MoS-VLA
\citep{zhao2025mosvla} learns a basis of skills during pretraining and infers
their combination from a new demonstration, while AtomicVLA
\citep{zhang2026atomicvla} organizes action generation around skill-guided
experts. DiTEA \citep{li2026ditea} and AdaMoE \citep{shen2025adamoe} place
sparsely activated experts inside the action model, with learned mechanisms
for expert selection and contribution. DyGRO-VLA \citep{lin2026dygrovla}
instead learns and routes grouped reinforcement-learning residuals for
cross-task policy optimization. These systems learn their modular structure or routing
mechanism as part of a prescribed training pipeline. \policyweave addresses a
different point in the workflow: task policies are post-trained independently
under a common action interface and are merged afterward.

Continual VLA learning updates one policy as tasks arrive sequentially. Recent
methods expand and route adapters, use replay, or continue supervised and
reinforcement learning while measuring retention of earlier skills
\citep{romer2026clare,liu2026resistant,liu2026lifelongrft,
zeng2026crlvla,hu2026simplerecipe}. This setting is related through its goal of
extending a deployed policy, but differs from post-hoc model merging because
the earlier and new skills participate in a sequential update process.

\subsection{VLA Policy Merging and Expert Selection}

Policy parameters participate in closed-loop control, so a small action change
can alter later observations and compound over a trajectory
\citep{ross2011dagger}. RETAIN \citep{yadav2026retain} merges a pretrained VLA
with one task-adapted descendant to preserve general capabilities during
robust finetuning. ReVLA \citep{dey2025revla} uses gradual model merging to
move OpenVLA's adapted visual backbones toward their pretrained states and
recover visual out-of-domain generalization. This is a within-policy
repair of visual representations rather than consolidation of independently
post-trained task experts. CORAL \citep{luo2026coral} and MergeVLA
\citep{fu2026mergevla} also maintain independently specialized components and
select among them at deployment without a supplied expert index.

\begin{table}[!t]
  \centering
  \caption{Comparison with CORAL and MergeVLA. ``Task unknown'' means
  that an expert index is not provided at inference time.}
  \label{tab:closest-vla-systems}
  \footnotesize
  \setlength{\tabcolsep}{3pt}
  \renewcommand{\arraystretch}{1.14}
  \begin{tabularx}{\linewidth}{@{}>{\raggedright\arraybackslash}p{0.15\linewidth}
    >{\raggedright\arraybackslash}X
    >{\raggedright\arraybackslash}X
    >{\raggedright\arraybackslash}X@{}}
    \toprule
    Aspect & CORAL & MergeVLA & \policyweave \\
    \midrule
    Expert construction
      & Isolated task LoRAs on a frozen base
      & Experts trained with a merge-oriented design
      & \cellcolor{policyweaveblue}Independent task LoRAs with a shared action interface \\
    \midrule
      Action model
      & Original action head retained and adapted by LoRA
      & Cross-attention-only expert; deeper task heads retained
      & \cellcolor{policyweaveblue}Original architecture retained; action encoder and decoder fixed \\
    \midrule
      Task-unknown selection
      & Externally specified instruction-to-expert function
      & Model-response-based task selection
      & \cellcolor{policyweaveblue}Expert-update responses with stability-adaptive selection \\
    \midrule
    Deployment
      & One selected LoRA folded into the base
      & Merged shared blocks with a selected mask and task head
      & \cellcolor{policyweaveblue}Static policy from the shared-interface bank, or context-guided sparse merging \\
    \midrule
    Static merging
      & Not part of the proposed method
      & Coupled with task-specific masks and heads
      & \cellcolor{policyweaveblue}Existing methods produce one policy without task-specific selection \\
    \bottomrule
  \end{tabularx}
\end{table}

\paragraph{CORAL.}
CORAL \citep{luo2026coral} freezes a pretrained VLA and learns an isolated
LoRA expert for each task. Its manager delegates expert choice to a client-side
instruction-to-expert function, then folds the selected adapter into the base
weights for standard inference. This design avoids
cross-task parameter interference by executing one isolated expert at a time.
The routing function is part of the deployment interface rather than a
specified model-derived scorer, and CORAL does not combine multiple task
updates within one deployed policy. In contrast, \policyweave derives its
decision from the VLA's context responses and can merge a sparse set of
compatible task updates.
To compare this deployment choice on a common expert bank,
Table~\ref{tab:robocasa-selection-baselines} evaluates two client-side hard selectors, a
trained context MLP and Semantic-NN, against PolicyWeave.

\paragraph{MergeVLA.}
MergeVLA \citep{fu2026mergevla} makes the VLA architecture merge-oriented. It applies task-specific
masks to merged VLM updates, replaces the action expert with
cross-attention-only blocks, and retains a task-specific deeper action head.
Its reported TA, WUDI, and TIES variants pair each
underlying operator with the task mask; applying standard merging methods directly to
the original VLA specialists gives zero success on LIBERO. When no task index
is supplied, its training-free router evaluates task-masked VLM variants and
selects the mask and action head whose representations best match
value-projection subspaces in the action expert. Thus, shared components are
merged, while the selected mask and action head remain task-specific at
execution.

\paragraph{Position of \policyweave.}
MergeVLA addresses mergeability through architecture design, whereas
\policyweave addresses it during independent post-training. Its shared
action interface allows existing static methods to produce a single policy
without the task-specific masks and heads used by MergeVLA. Expert
compatibility can thus be studied separately from the merging algorithm.
For context-guided sparse merging, \policyweave specifies both the
expert-response scorer and stability-based set selection, in contrast to
CORAL's externally specified single-expert selection.

\section{Additional PolicyWeave Details}
\label{app:contextvote-details}

\subsection{Shared Action Interface}
\label{app:shared-interface-analysis}
We use a scalar linear policy to isolate the effect of a varying action
interface. Let $z$ be a common input, $h_t$ expert $t$'s hidden-layer
coefficient, and $d_t$ its output-interface coefficient. Its physical action
is $a_t=d_t h_t z$. For $K$ experts, define the parameter averages
$\bar d=\frac1K\sum_{t=1}^K d_t$ and
$\bar h=\frac1K\sum_{t=1}^K h_t$. The parameter-averaged policy produces
$\bar a=\bar d\bar h z$. Its difference from the average expert action is
\begin{equation}
  \bar a-\frac1K\sum_{t=1}^K a_t
  =-\frac1K\sum_{t=1}^K(d_t-\bar d)(h_t-\bar h)z.
  \label{eq:interface-averaging-discrepancy}
\end{equation}
Expanding the centered product gives
$\frac1K\sum_t(d_t-\bar d)(h_t-\bar h)
=\frac1K\sum_t d_t h_t-\bar d\bar h$, establishing the identity.
Thus, joint variation in the interface and hidden-layer coefficients can
make parameter averaging differ from action averaging. When all experts
share $d_t=d_0$, the right-hand side is zero and
$\bar a=d_0\bar h z=\frac1K\sum_t a_t$.

For a concrete example, take $(d_1,h_1)=(1,1)$ and
$(d_2,h_2)=(2,\tfrac12)$. Both experts produce
$a_t=z$. Averaging their parameters instead gives
\begin{equation}
  \bar a
  =\left(\frac{d_1+d_2}{2}\right)
   \left(\frac{h_1+h_2}{2}\right)z
  =\frac{9}{8}z.
  \label{eq:interface-averaging-example}
\end{equation}
The second expert halves its hidden prediction and doubles the output scale,
preserving its action before merging. Averaging both coefficients breaks
this compensation. A shared interface eliminates this discrepancy in the
linear model; whether averaging different task behaviors yields useful
control remains a separate question.

\subsection{Context-Response Scoring}
\label{app:contextvote-efficiency}
\paragraph{Score interpretation.}
For one expert and scoring layer, write
$\mathbf U=\Delta\mathbf W^{\mathrm K}_{t,\ell}$ and
$\mathbf H=\mathbf H_\ell$, and define
$\mathbf Q=\mathbf U^\top\mathbf U$ and
$\mathbf S=\mathbf H\mathbf H^\top$.
Both are positive semidefinite matrices in
$\mathbb R^{d_{\mathrm{in}}\times d_{\mathrm{in}}}$.
The matrix $\mathbf Q$ describes the input directions to which the update
responds, while $\mathbf S$ is the unnormalized second-moment matrix of the
context tokens. For nonzero $\mathbf U$ and $\mathbf H$, cyclicity of the
trace gives
\begin{equation}
  e_{t,\ell}
  =\frac{\operatorname{tr}(\mathbf U\mathbf H\mathbf H^\top\mathbf U^\top)}
  {\operatorname{tr}(\mathbf U^\top\mathbf U)
   \operatorname{tr}(\mathbf H\mathbf H^\top)}
  =\frac{\operatorname{tr}(\mathbf Q\mathbf S)}
  {\operatorname{tr}(\mathbf Q)\operatorname{tr}(\mathbf S)}.
  \label{eq:context-response-trace}
\end{equation}
Thus, the score measures directional overlap between the update's sensitivity
and the context's second moment. Since
$\operatorname{tr}(\mathbf Q\mathbf S)\geq0$ and
$\operatorname{tr}(\mathbf Q\mathbf S)
\leq\lambda_{\max}(\mathbf Q)\operatorname{tr}(\mathbf S)
\leq\operatorname{tr}(\mathbf Q)\operatorname{tr}(\mathbf S)$,
where $\lambda_{\max}$ denotes the largest eigenvalue, we have
$0\leq e_{t,\ell}\leq1$.
Multiplying either $\mathbf U$ or $\mathbf H$ by any nonzero scalar leaves
the score unchanged. This removes overall magnitude as a source of score
differences; expert identification depends on how the learned update
directions align with the observed context.

\paragraph{Tokenwise form.}
In Equation~\eqref{eq:contextvote-score}, $\mathbf{H}_\ell$ stacks only valid
context-token representations. For a padded sequence of $P$ tokens, let
$\mathbf{h}_{\ell,p}$ be the input of token $p$ to the key projection and let
$\mu_p\in\{0,1\}$ indicate whether that token is valid. The equivalent
tokenwise score is
\begin{equation}
  e_{t,\ell}=
  \frac{\sum_{p=1}^{P}\mu_p
    \lVert\Delta\mathbf{W}^{\mathrm{K}}_{t,\ell}\mathbf{h}_{\ell,p}\rVert_2^2}
  {\lVert\Delta\mathbf{W}^{\mathrm{K}}_{t,\ell}\rVert_F^2
    \sum_{p=1}^{P}\mu_p\lVert\mathbf{h}_{\ell,p}\rVert_2^2}.
  \label{eq:contextvote-score-tokenwise}
\end{equation}
The key-projection update includes all attention heads. Summing squared
column norms gives the squared matrix norm in the main-text expression;
padding tokens contribute to neither the numerator nor the denominator.

\paragraph{Efficient low-rank computation.}
The response score in Equation~\eqref{eq:contextvote-score} can be evaluated
without materializing the dense update $\Delta\mathbf{W}^{\mathrm{K}}_{t,\ell}$.
The LoRA parameterization at an adapted layer is
\begin{equation}
  \Delta\mathbf{W}_{t,\ell}
  =\frac{\alpha_\ell}{r_\ell}\mathbf{B}_{t,\ell}\mathbf{A}_{t,\ell},
  \label{eq:lora-update}
\end{equation}
where $r_\ell$ is the rank, $\alpha_\ell$ is the scaling factor, and
$\mathbf{A}_{t,\ell}\in\mathbb{R}^{r_\ell\times d_{\mathrm{in}}}$ and
$\mathbf{B}_{t,\ell}\in\mathbb{R}^{d_{\mathrm{out}}\times r_\ell}$ are the
LoRA factors for layer input and output dimensions $d_{\mathrm{in}}$ and
$d_{\mathrm{out}}$. The response to a context token $\mathbf{h}$ satisfies
\begin{equation}
  \left\lVert \Delta\mathbf{W}_{t,\ell}\mathbf{h}\right\rVert_2^2
  =
  \left(\frac{\alpha_\ell}{r_\ell}\right)^2
  (\mathbf{A}_{t,\ell}\mathbf{h})^\top
  (\mathbf{B}_{t,\ell}^\top \mathbf{B}_{t,\ell})
  (\mathbf{A}_{t,\ell}\mathbf{h}),
  \label{eq:contextvote-factor-response}
\end{equation}
while the update norm is
\begin{equation}
  \left\lVert \Delta\mathbf{W}_{t,\ell}\right\rVert_F^2
  =
  \left(\frac{\alpha_\ell}{r_\ell}\right)^2
  \operatorname{tr}\!\left(
    \mathbf{A}_{t,\ell}\mathbf{A}_{t,\ell}^\top
    \mathbf{B}_{t,\ell}^\top \mathbf{B}_{t,\ell}
  \right).
  \label{eq:contextvote-factor-norm}
\end{equation}
The LoRA scaling cancels in the normalized response. A single base-context
pass therefore suffices to score all experts using low-rank products.
With $\mathbf{B}_{t,\ell}^{\top}\mathbf{B}_{t,\ell}$ and update
norms cached, scoring one expert at one layer costs
$\mathcal{O}\!\left(P(r_\ell d_{\mathrm{in}}+r_\ell^2)\right)$, so the full
scorer grows linearly with the number of experts and scoring layers.

\subsection{Stability of Expert Selection}
\label{app:selection-stability}
The single-layer removal test also controls sensitivity to partial reductions
in layer contributions. We analyze the score ranking for a fixed context.

\paragraph{Proposition: stability under layer downweighting.}
Let $a=\rho_1$ be the unique highest-scoring expert. Suppose $a$ remains the
unique highest-scoring expert after removing any one scoring layer.
For any layer-weight reductions $\delta_\ell\geq0$ satisfying
$\sum_{\ell\in\mathcal C}\delta_\ell\leq1$, define
\begin{equation}
  \widetilde{s}_t
  =\sum_{\ell\in\mathcal C}(1-\delta_\ell)e_{t,\ell}.
  \label{eq:selection-downweighted-score}
\end{equation}
Then $a$ remains the unique highest-scoring expert under
$\widetilde{s}_t$. Thus, the test covers not only removal of one
layer, but also partial downweighting spread across several layers, such as
halving the contributions of two layers.

\paragraph{Proof.}
For a competing expert $j\ne a$, let $m_j=s_a-s_j$ denote the original
score margin. After removing layer $\ell$, this margin is
$m_j-(e_{a,\ell}-e_{j,\ell})$. The assumed strict stability is therefore
equivalent to
\begin{equation}
  m_j>\max_{\ell\in\mathcal C}(e_{a,\ell}-e_{j,\ell}),
  \qquad \text{for every }j\ne a.
  \label{eq:selection-stability-margin}
\end{equation}
Write $D=\sum_{\ell\in\mathcal C}\delta_\ell$. Since
$s_t^{(-\ell)}=s_t-e_{t,\ell}$, the downweighted scores satisfy
\begin{equation}
  \widetilde{s}_t
  =(1-D)s_t+\sum_{\ell\in\mathcal C}\delta_\ell s_t^{(-\ell)}.
  \label{eq:selection-score-convex-combination}
\end{equation}
The coefficients are nonnegative and sum to one. Expert $a$ has a strictly
positive margin over every competitor in each score vector on the right,
so it also has a strictly positive margin in their convex combination.
Hence the leading expert is unchanged.

\paragraph{Adaptive set size.}
With a fixed tie-breaking rule, Equation~\eqref{eq:contextvote-adaptive-set}
returns the smallest prefix of the aggregate ranking that contains every
leave-one-layer-out top-ranked expert. Any shorter prefix excludes an expert
whose rank attains $k^\star$, whereas the prefix of length $k^\star$ contains
them all. In particular, $k^\star=1$ exactly when the original leading expert
remains top-ranked under every layer removal; in this case,
Equation~\eqref{eq:contextvote-compose} recovers that expert's parameter update
exactly.

\subsection{Episode-Level Inference}
\label{app:contextvote-inference}
Algorithm~\ref{alg:contextvote-inference} gives the complete inference
procedure. Here $\mathbf{x}_{\tau}$ contains the observation and instruction
at control step $\tau$, and $\widehat{\mathbf{a}}_{\tau}$ is the predicted
action chunk.
We use $\policy_{\theta,m_0}$ for the policy with parameters $\theta$ and the
shared action normalization metadata $m_0$. Encoding $\mathbf{x}_0$ supplies
the initial visual-language context $c$ used in
Equation~\eqref{eq:contextvote-compose}.
\begin{algorithm}[!t]
  \caption{Episode-level inference with \policyweave}
  \label{alg:contextvote-inference}
  \begin{algorithmic}[1]
    \Require Base policy $\policy_{\baseparams,m_0}$; PolicyWeave updates
    $\{\taskdelta{t}\}_{t\in\taskset}$; initial context
    $\mathbf{x}_0$; scoring layers $\mathcal{C}$; adapted layers $\mathcal{H}$
    \State Disable all task adapters and encode $\mathbf{x}_0$ with the base
    policy to obtain the valid key-projection inputs
    $\{\mathbf{H}_{\ell}\}_{\ell\in\mathcal{C}}$
    \ForAll{$(t,\ell)\in\taskset\times\mathcal{C}$}
      \State Compute $e_{t,\ell}$ from Equation~\eqref{eq:contextvote-score},
      using Equations~\eqref{eq:contextvote-factor-response}
      and~\eqref{eq:contextvote-factor-norm}
    \EndFor
    \State $s_t \gets \sum_{\ell\in\mathcal{C}} e_{t,\ell}$ for every
    $t\in\taskset$
    \State Order the experts as $\rho=(\rho_1,\ldots,\rho_K)$ such that
    $s_{\rho_1}\geq\cdots\geq s_{\rho_K}$
    \ForAll{$\ell\in\mathcal{C}$}
      \State $t_\ell^\star \gets \arg\max_{t\in\taskset}(s_t-e_{t,\ell})$
    \EndFor
    \State $k^\star\gets
    \max_{\ell\in\mathcal{C}}\operatorname{rank}_{\rho}(t_\ell^\star)$
    \State $\mathcal{I}^\star\gets\{\rho_1,\ldots,\rho_{k^\star}\}$
    \State $\gamma_t\gets s_t/\sum_{j\in\mathcal{I}^\star}s_j$ for every
    $t\in\mathcal{I}^\star$
    \ForAll{$u\in\mathcal{H}$}
      \State $\Delta \mathbf{W}_u \gets
      \sum_{t\in\mathcal{I}^\star}\gamma_t\Delta \mathbf{W}_{t,u}$
    \EndFor
    \State Add $\{\Delta \mathbf{W}_u\}_{u\in\mathcal{H}}$ to the base weights,
    keeping the shared action interface fixed
    \State $\tau\gets 0$
    \While{the episode has not terminated}
      \State Predict $\widehat{\mathbf{a}}_{\tau}$ from
      $\mathbf{x}_{\tau}$ with the merged policy
      \State Execute $\widehat{\mathbf{a}}_{\tau}$ and observe
      $\mathbf{x}_{\tau+1}$
      \State $\tau\gets\tau+1$
    \EndWhile
  \end{algorithmic}
\end{algorithm}
The base pass supplies routing scores; the merged update is then applied to
all 120 adapted action-model modules and held fixed for the episode.
In practice, we apply the merged update in low-rank form without
materializing a dense weight matrix.
Hard Top-1 and Hard Top-2 use the same expert scores and score-normalized
weights but fix the selected set size.

\section{Experimental Setup and Protocols}
\label{app:experimental-details}

\subsection{Expert Banks}
\label{app:expert-banks}
For RoboCasa365 \citep{nasiriany2026robocasa365}, experts start from the
\href{https://huggingface.co/robocasa/robocasa365_checkpoints/tree/main/gr00t_n1-5/foundation_model_learning/pretraining/checkpoint-80000}{official GR00T N1.5 checkpoint}
\citep{bjorck2025groot} trained on the benchmark's pretraining split.
LIBERO-10 experts start from a
\href{https://huggingface.co/youliangtan/gr00t-n1.5-libero-90-posttrain}{public GR00T N1.5 checkpoint}
trained on LIBERO-90.
Within each expert bank, every checkpoint is derived from the same base model
and retains the shared \policyweave action normalization, encoder, and decoder.
Only rank-8 LoRA adapters on the 120 action-model linear modules are updated.
RoboCasa365 contains 18 SFT experts, each trained with 50 demonstrations, which
corresponds to 10\% of the available training data. Its second bank continues
each SFT expert with Flow-SDE reinforcement learning
\citep{chen2025pirl}. Thus, every RL-adapted expert is initialized from its
corresponding SFT checkpoint.
LIBERO-10 contains a single bank of 10 SFT experts trained with all available
demonstrations for their respective tasks. The number of demonstrations ranges
from 29 to 49, with a mean of 37.9 per task.

\subsection{Training Details}
\label{app:training-details}
Each task-specific job uses one NVIDIA RTX PRO 6000 Blackwell GPU. SFT uses the
flow-matching objective in bfloat16, AdamW with $(\beta_1,\beta_2)=(0.95,0.999)$
and $\epsilon=10^{-8}$, a cosine schedule with 5\% warmup, and no gradient
accumulation. Both benchmarks use LoRA rank 8, scaling factor 16, and dropout
0; Table~\ref{tab:sft-training-details} gives the remaining settings.

\begin{table}[!ht]
  \centering
  \caption{SFT configuration for each task-specific expert.}
  \label{tab:sft-training-details}
  \setlength{\tabcolsep}{3pt}
  \begin{tabular*}{\linewidth}{@{\extracolsep{\fill}}lccccc@{}}
    \toprule
    Benchmark & Data per task & Steps & Batch & LR & Weight decay \\
    \midrule
    RoboCasa365 & 50 demos (10\%) & 10,000 & 128 & $3\!\times\!10^{-5}$ & $10^{-5}$ \\
    LIBERO-10 & all (37.9 avg.) & 10,000 & 128 & $3\!\times\!10^{-5}$ & $10^{-5}$ \\
    \bottomrule
  \end{tabular*}
\end{table}

\paragraph{Co-training baselines.}
The RoboCasa365 \textit{Co-training} SFT baseline uses the official checkpoint
co-trained on all 18 target tasks with 10\% of the training data
\citep{nasiriany2026robocasa365}. We initialize our joint RL continuation from
this checkpoint. For LIBERO-10, we re-evaluate the GR00T N1.5 co-training
checkpoint provided in the
\href{https://github.com/NVIDIA/Isaac-GR00T/blob/n1.5-release/examples/Libero/README.md}{official LIBERO example}
of NVIDIA's Isaac-GR00T repository (\texttt{n1.5-release}).

\paragraph{Reinforcement learning.}
RoboCasa365 RL uses PPO actor-critic optimization with Flow-SDE action
log-probabilities, four rollout epochs, two update epochs, $\gamma=0.99$, GAE
$0.95$, policy and value clipping at $0.2$, weight decay $0.01$, and
gradient-norm clipping at $10$. LoRA optimization uses float32 and policy
forward passes use bfloat16. The selected task-specific RL checkpoints sum to 610 steps;
the \textit{Co-training} SFT policy undergoes 600 joint RL steps.
Task-specific checkpoints are selected by per-task validation success, while
the cotrained checkpoint is selected by mean validation success across the
18 tasks. Checkpoint selection and final evaluation use separate rollout sets.
Tables~\ref{tab:rl-training-details} and~\ref{tab:multitask-rl-curve}
give the configurations and validation curve; the selected joint checkpoint
is at step 600.

\begin{table}[!ht]
  \centering
  \caption{RoboCasa365 SFT then RL configurations at the selected checkpoints.
  M denotes million; batch sizes are micro/global.}
  \label{tab:rl-training-details}
  \setlength{\tabcolsep}{3pt}
  \begin{tabular*}{\linewidth}{@{\extracolsep{\fill}}lcccccc@{}}
    \toprule
    Training & RL steps & Envs & Env. steps (M) & Batch & Actor LR & Value LR \\
    \midrule
    Individual & 33.9 avg. (610 total) & 16 & 23.74 & 8/304 & $10^{-5}$ & $10^{-4}$ \\
    Co-training & 600 total & 18 & 26.27 & 8/304 & $10^{-5}$ & $10^{-4}$ \\
    \bottomrule
  \end{tabular*}
\end{table}

\begin{table}[!ht]
  \centering
  \caption{RoboCasa365 co-training RL checkpoint-selection curve. Values are
  mean validation success rates (\%) over the 18 tasks.}
  \label{tab:multitask-rl-curve}
  \setlength{\tabcolsep}{5pt}
  \begin{tabular*}{\linewidth}{@{\extracolsep{\fill}}lrrrrrrr@{}}
    \toprule
    RL step & 0 & 100 & 200 & 300 & 400 & 500 & 600 \\
    \midrule
    Success & 58.1 & 57.2 & 59.8 & 59.6 & 60.1 & 59.8 & \textbf{60.7} \\
    \bottomrule
  \end{tabular*}
\end{table}

\subsection{Rollout Protocol and Evaluation Metrics}
\label{app:rollout-protocol}
RoboCasa365 evaluation uses the official version 1.0.1 target split, four Euler
integration steps for flow sampling, and an action horizon of 16. LIBERO-10
episodes contain at most 480 simulator steps and execute action chunks of length
10. These settings are held fixed across the methods compared within each
benchmark. RoboCasa365 results are averaged over 50 episodes per task.
On LIBERO-10, each method is evaluated in separate closed-loop rollouts
using the same reset schedule and policy-noise seeds.

For LIBERO-10, let $\mathcal{G}_e$ be the set of goal predicates for episode
$e$, and let $y_{e,\tau,j}$ indicate whether predicate $j$ is satisfied at
timestep $\tau$. We define partial success as
\begin{equation}
  \operatorname{PS}
  =\frac{1}{N_{\mathrm{eval}}}
  \sum_{e=1}^{N_{\mathrm{eval}}}
  \max_{1\leq\tau\leq T_e}
  \frac{1}{\lvert\mathcal{G}_e\rvert}
  \sum_{j\in\mathcal{G}_e}y_{e,\tau,j},
  \label{eq:partial-success}
\end{equation}
where $T_e$ is the episode length and $N_{\mathrm{eval}}$ is the number of
evaluation episodes. This metric measures peak simultaneous goal completion,
capturing partial progress on long-horizon tasks.

\subsection{Merging Baselines}
\label{app:merging-baselines}
All merging methods operate on the induced weight updates
$\Delta \mathbf{W}_{t,u}$. WA assigns a coefficient of $1/K$ to each of the
$K$ task updates. TIES, TSV-Merge, and ISO-CTS operate in the shared low-rank
Core Space \citep{panariello2025corespace} before reconstruction in weight
space. SSR-Merge uses its LoRA subspace signal-routing formulation.
Let $\lambda$ denote the global
coefficient applied to the merged task update. For TIES, TSV-Merge, and ISO-CTS,
we select $\lambda\in\{0.1,0.2,\ldots,1.0\}$ using validation rollouts across
all target tasks, with seeds disjoint from the final evaluation, and hold it
fixed during testing.
The selected values are $0.8$ for TIES, $0.3$ for TSV-Merge, and $0.4$ for
ISO-CTS across all three expert banks. We set $\lambda=1.0$ for SSR-Merge. All other hyperparameters
follow the defaults of the corresponding methods
\citep{yadav2023ties,gargiulo2025tsv,marczak2025isotropic,wei2026ssrmerge}.
For the RoboCasa365 SSR-Merge baseline, the routing transform is estimated from
five successful trajectories per task, with ten states sampled from each
trajectory. In the main comparisons, all static merged checkpoints are
materialized on the same base policy and retain the shared action interface.

\subsection{Expert Selection Baselines}
\label{app:selection-baseline-setup}
\paragraph{Trainable MLP.}
The MLP learns a context-to-expert mapping, analogous to the learned routing
in Twin-Merging \citep{lu2024twinmerging}, and selects one expert.
On both benchmarks, it receives a masked mean of the 2,048-dimensional frozen-base
visual-language context and has two hidden layers of width 128. Its output
contains one score per expert: 18 on RoboCasa365 and 10 on LIBERO-10.
The highest-scoring expert is selected for the episode.
We train it on 10 demonstrations per task, using balanced cross-entropy and
500 AdamW steps with batch size 2 and learning rate $10^{-3}$ on both benchmarks.
We use the final training checkpoint.


\paragraph{Semantic-NN.}
Semantic-NN (Sem-NN) uses the frozen \texttt{all-mpnet-base-v2} text encoder
\citep{reimers2019sentencebert,song2020mpnet}. Each expert is represented by
one lowercased, word-separated task-name prototype. Selection uses cosine
similarity to the instruction embedding, without demonstration calibration.

\subsection{Published Leaderboard References}
\label{app:leaderboard-references}
Figure~\ref{fig:robocasa-leaderboard} includes selected results from the
official RoboCasa365 leaderboard.
The selected Atomic-Seen scores cover 18 tasks in pretraining kitchens under
the multi-task learning protocol. Our results use target kitchens under the
foundation-model learning protocol, with 50 SFT demonstrations per task
(10\% of target demonstrations) followed by RL. This fraction refers to SFT
data, separately from base pretraining and RL interaction. Published entries
use their original configurations; the leaderboard's GR00T N1.5 entry uses
RoboCasa 1.0.1 with a rollout horizon 1.5 times that of the original paper.

\section{Additional Results and Analysis}
\label{app:additional-results}

\subsection{Action-Interface Compatibility}
\label{app:mergeability-metrics}
\paragraph{Task-specific action-interface baseline.}
This rank-8 LoRA bank uses the same GR00T N1.5 base and 18-task evaluation
protocol. It additionally adapts the action encoder and decoder and stores
task-specific normalization. Each merging method combines the learned action modules,
while evaluation manually loads the target task's normalization for the merged
checkpoint and thus assumes task identity.
\begin{table}[!ht]
  \centering
  \caption{Static merging under task-specific and shared action interfaces
  on RoboCasa365. Retention is the success rate of
  each merging method divided by
  Individual success. All entries are percentages.}
  \label{tab:action-interface-static-methods}
  \setlength{\tabcolsep}{3pt}
  \begin{tabularx}{\linewidth}{@{}l*{6}{>{\centering\arraybackslash}X}@{}}
    \toprule
    Action interface & Individual & WA & TIES & ISO-CTS & TSV-Merge &
    \shortstack{Retention\\range} \\
    \midrule
    Task-specific & 64.1 & 16.1 & 13.9 & 18.9 & 18.9 & 21.7--29.5 \\
    \rowcolor{policyweaveblue}\cellcolor{white}\policyweave
    & 65.6 & 49.6 & 52.9 & 54.1 & 54.7 & 75.6--83.4 \\
    \bottomrule
  \end{tabularx}
\end{table}

\paragraph{LIBERO-10 without a shared action interface.}
For the \policyweave variant in Table~\ref{tab:libero-selection-main},
single-expert selection uses the selected expert's normalization, encoder,
and decoder. When multiple experts are selected, their normalization
statistics are combined, and their encoder and decoder parameters are merged
with the same weights as their hidden-layer updates. The interface therefore
follows the selected experts rather than a supplied task identity.
Table~\ref{tab:libero-no-shared-interface} reports additional aggregate results
for the four static merging methods without a shared action interface.
All four methods have zero final success and 8.3--13.2\% partial success
(Equation~\eqref{eq:partial-success}), showing limited intermediate-goal progress.
\begin{table}[!ht]
  \centering
  \caption{LIBERO-10 aggregate success rates without a shared action interface.
  Partial success measures the maximum fraction of goal predicates satisfied
  simultaneously during an episode, averaged across episodes.
  All entries are percentages.}
  \label{tab:libero-no-shared-interface}
  \setlength{\tabcolsep}{6pt}
  \begin{tabularx}{\linewidth}{@{}l*{4}{>{\centering\arraybackslash}X}@{}}
    \toprule
    Metric & WA & TIES & ISO-CTS & TSV-Merge \\
    \midrule
    Final success & 0.0 & 0.0 & 0.0 & 0.0 \\
    Partial success & 10.6 & 9.8 & 8.3 & 13.2 \\
    \bottomrule
  \end{tabularx}
\end{table}

\paragraph{Normalization and closed-loop diagnostics.}
For the paired normalization diagnostic, relative $\ell_2$ shift is
$\lVert \mathbf{A}_{\mathrm{shared}}-\mathbf{A}_{\mathrm{task}}\rVert_F /
\lVert \mathbf{A}_{\mathrm{task}}\rVert_F$, where each matrix contains the same
sampled raw actions transformed by the indicated normalization. We use 1,800
raw actions from 18 RoboCasa365 tasks, sampling 20 timesteps from five
demonstrations per task.
The heatmap averages absolute differences in normalized actions over samples
and dimensions within each action group.

Figure~\ref{fig:p92-coordinate-geometry}(c) compares a task expert,
task-specific WA, and shared-interface WA across three matched scenes, with
paired denoising noise. Rollouts use a 600-step horizon, action chunks of
length 16, and four Euler steps. Base displacement is
$100\lVert\mathbf{p}^{xy}_\tau-\mathbf{p}^{xy}_0\rVert_2$ in centimetres, where
$\mathbf{p}^{xy}_\tau$ is the mobile base's planar world position in metres
at timestep $\tau$. The figure shows the first scene
at step 160, with the expert trajectory as a visual reference.
Across the three scenes, maximum base displacement is 88.3--122.0\,cm for
task-specific WA and 0.6--1.6\,cm for shared-interface WA.

\subsection{LIBERO-10 Per-Task Results}
\label{app:libero-results}
\paragraph{Published co-training references.}
Figure~\ref{fig:libero-main} uses published LIBERO-Long success rates for
jointly fine-tuned OpenVLA \citep{kim2025openvla}, SpatialVLA
\citep{qu2025spatialvla}, and CoT-VLA \citep{zhao2025cotvla}, under their
original configurations. These reports provide final success only.
The GR00T N1.5 result is our re-evaluation of the official checkpoint described in
Appendix~\ref{app:training-details}.

\paragraph{Task-level results.}
Table~\ref{tab:libero-main} reports final and partial success for each task
underlying the averages in Figure~\ref{fig:libero-main}. Each task is evaluated
over 50 episodes; the Average row gives the equally weighted mean across the
10 tasks. Partial success is defined in Equation~\ref{eq:partial-success}.
\begin{table}[!ht]
\caption{LIBERO-10 success (\%), averaged over 50 episodes per task. Each cell
reports final success above \textcolor{black!60}{partial success}, the maximum
fraction of goal predicates satisfied simultaneously during an episode.
 Bold marks the
best merging result for each task and metric.}
\label{tab:libero-main}
\centering
\small
\setlength{\tabcolsep}{2.5pt}
\renewcommand{\arraystretch}{1.02}
\newcommand{\liberotask}[1]{\mbox{\textsc{#1}}}
\newcommand{\liberoheader}[1]{\makebox[\linewidth][c]{#1}}
\newcommand{\liberopair}[2]{\makebox[\linewidth][c]{%
  \shortstack[c]{#1\\[-1pt]\textcolor{black!60}{#2}}}}
\newcommand{\liberobestpair}[2]{\cellcolor{policyweaveblue}\makebox[\linewidth][c]{%
  \shortstack[c]{\textbf{#1}\\[-1pt]\textcolor{black!60}{\textbf{#2}}}}}
\newcommand{\liberoindividualpair}[2]{\cellcolor{gray!15}\makebox[\linewidth][c]{%
  \shortstack[c]{#1\\[-1pt]\textcolor{black!60}{#2}}}}
\begin{tabularx}{\textwidth}{@{}p{0.46\textwidth}*{7}{>{\centering\arraybackslash}X}@{}}
\toprule
& \multicolumn{2}{c}{Baselines}
& \multicolumn{5}{c}{Merging methods} \\
\cmidrule(lr){2-3}\cmidrule(lr){4-8}
Task &
\liberoheader{\textit{Indiv.}} &
\liberoheader{\shortstack[c]{GR00T\\[-1pt]N1.5}} &
\liberoheader{WA} &
\liberoheader{TIES} &
\liberoheader{\shortstack[c]{ISO-\\[-1pt]CTS}} &
\liberoheader{TSV} &
\liberoheader{\shortstack[c]{\textbf{Policy}\\[-1pt]\textbf{Weave}}} \\
\midrule
\liberotask{soup + sauce $\rightarrow$ basket}
  & \liberoindividualpair{88}{91} & \liberopair{72}{80} & \liberopair{24}{62} & \liberopair{34}{67} & \liberopair{28}{63} & \liberopair{48}{73} & \liberobestpair{84}{89} \\
\liberotask{cream cheese + butter $\rightarrow$ basket}
  & \liberoindividualpair{86}{92} & \liberopair{84}{92} & \liberopair{10}{51} & \liberopair{12}{52} & \liberopair{24}{58} & \liberopair{34}{65} & \liberobestpair{70}{84} \\
\liberotask{turn on stove + moka pot}
  & \liberoindividualpair{94}{96} & \liberopair{78}{88} & \liberopair{0}{50} & \liberopair{0}{50} & \liberopair{0}{50} & \liberopair{0}{50} & \liberobestpair{90}{94} \\
\liberotask{black bowl $\rightarrow$ bottom drawer + close}
  & \liberoindividualpair{90}{95} & \liberopair{90}{94} & \liberopair{0}{50} & \liberopair{2}{51} & \liberopair{2}{51} & \liberopair{18}{58} & \liberobestpair{74}{87} \\
\liberotask{two mugs $\rightarrow$ two plates}
  & \liberoindividualpair{66}{79} & \liberopair{82}{89} & \liberopair{6}{49} & \liberopair{2}{46} & \liberopair{10}{50} & \liberopair{58}{76} & \liberobestpair{66}{78} \\
\liberotask{book $\rightarrow$ caddy back compartment}
  & \liberoindividualpair{98}{98} & \liberopair{100}{100} & \liberopair{92}{92} & \liberopair{90}{90} & \liberopair{92}{92} & \liberopair{92}{92} & \liberobestpair{100}{100} \\
\liberotask{mug $\rightarrow$ plate + pudding right}
  & \liberoindividualpair{88}{93} & \liberopair{64}{79} & \liberopair{2}{45} & \liberopair{4}{49} & \liberopair{2}{49} & \liberopair{10}{48} & \liberobestpair{78}{88} \\
\liberotask{soup + cream cheese $\rightarrow$ basket}
  & \liberoindividualpair{88}{91} & \liberopair{68}{79} & \liberopair{6}{49} & \liberopair{14}{52} & \liberopair{8}{54} & \liberopair{18}{59} & \liberobestpair{88}{92} \\
\liberotask{two moka pots $\rightarrow$ stove}
  & \liberoindividualpair{32}{70} & \liberopair{14}{62} & \liberopair{0}{65} & \liberopair{2}{62} & \liberopair{0}{65} & \liberopair{4}{64} & \liberobestpair{32}{73} \\
\liberotask{mug $\rightarrow$ microwave + close}
  & \liberoindividualpair{78}{84} & \liberopair{72}{78} & \liberopair{0}{26} & \liberopair{0}{31} & \liberopair{0}{24} & \liberopair{0}{27} & \liberobestpair{72}{79} \\
\addlinespace[1pt]
\textbf{Average}
  & \liberoindividualpair{80.8}{88.9} & \liberopair{72.4}{84.1} & \liberopair{14.0}{53.9} & \liberopair{16.0}{55.0} & \liberopair{16.6}{55.6} & \liberopair{28.2}{61.2} & \liberobestpair{75.4}{86.4} \\
\bottomrule
\end{tabularx}
\end{table}

\paragraph{MergeVLA reproduction.}
We reproduce MergeVLA \citep{fu2026mergevla} on LIBERO-10 using its own
backbone and fine-tuning pipeline. Table~\ref{tab:mergevla-reproduction}
reports its individual and merged results.
\begin{table}[!ht]
  \centering
  \caption{MergeVLA reproduction on LIBERO-10. FT evaluates the task-specific
  fine-tuned policies and Merged evaluates the resulting merged policy.}
  \label{tab:mergevla-reproduction}
  \setlength{\tabcolsep}{14pt}
  \renewcommand{\arraystretch}{0.95}
  \begin{tabular*}{\linewidth}{@{\extracolsep{\fill}}lrr@{}}
    \toprule
    Task index & FT (\%) & Merged (\%) \\
    \midrule
    0 & 86 & 76 \\
    1 & 98 & 98 \\
    2 & 100 & 94 \\
    3 & 100 & 74 \\
    4 & 84 & 78 \\
    5 & 90 & 82 \\
    6 & 80 & 30 \\
    7 & 94 & 68 \\
    8 & 72 & 32 \\
    9 & 92 & 76 \\
    \midrule
    \textbf{Average} & \textbf{89.6} & \textbf{70.8} \\
    \bottomrule
  \end{tabular*}
\end{table}

\subsection{Expert Selection Statistics}
\label{app:routing-selection-statistics}

Table~\ref{tab:selection-size-distribution} reports the selection-size
distributions for the shared-interface expert banks.
\begin{table}[!t]
  \centering
  \caption{Expert selection sizes under \policyweave. Columns $k=1$ through
  $k=4$ give episode counts; $k>1$ gives the percentage of episodes selecting
  multiple experts. No episode selects more than four experts.}
  \label{tab:selection-size-distribution}
  \small
  \setlength{\tabcolsep}{4pt}
  \renewcommand{\arraystretch}{1.08}
  \begin{tabular*}{\linewidth}{@{\extracolsep{\fill}}lrrrrrrr@{}}
    \toprule
    Expert bank & Episodes & $k=1$ & $k=2$ & $k=3$ & $k=4$ & $k>1$ (\%) & Mean $k$ \\
    \midrule
    RoboCasa365 SFT & 900 & 887 & 13 & 0 & 0 & 1.44 & 1.014 \\
    RoboCasa365 SFT then RL & 900 & 889 & 11 & 0 & 0 & 1.22 & 1.012 \\
    LIBERO-10 SFT & 500 & 467 & 29 & 3 & 1 & 6.60 & 1.076 \\
    \bottomrule
  \end{tabular*}
\end{table}

\begin{figure}[!t]
  \centering
  \begin{minipage}[t]{0.49\linewidth}
    \centering
    \textbf{(a) RoboCasa365 SFT}\\[4pt]
    \includegraphics[width=\linewidth,trim=0 0 0 21,clip]{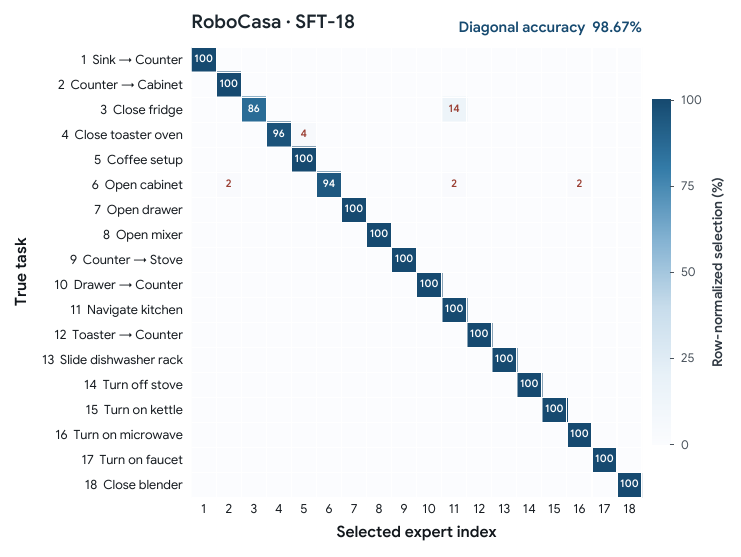}
  \end{minipage}\hfill
  \begin{minipage}[t]{0.49\linewidth}
    \centering
    \textbf{(b) LIBERO-10 SFT}\\[4pt]
    \includegraphics[width=\linewidth]{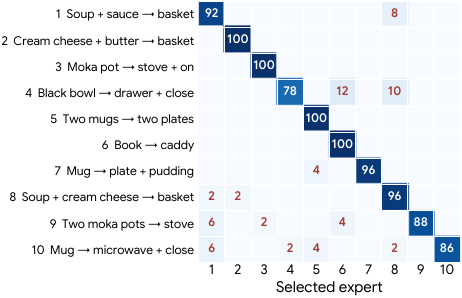}
  \end{minipage}
  \caption{Row-normalized Top-1 routing confusion matrices (\%) for the SFT banks:
  (a) RoboCasa365 (900 episodes); (b) LIBERO-10 (500 episodes).
  Rows indicate the requested task; columns indicate the selected expert.}
  \label{fig:libero-routing-confusion}
\end{figure}

Figure~\ref{fig:libero-routing-confusion} compares expert selection on the
RoboCasa365 and LIBERO-10 SFT banks. RoboCasa365 is nearly diagonal, whereas
LIBERO-10 includes confusions among tasks sharing objects or intermediate
goals, such as basket placement and manipulation of mugs or moka pots.

\subsection{Instruction and Observation Robustness}
\label{app:router-robustness}
The instruction comparison in Table~\ref{tab:robocasa-selection-baselines} uses
the RoboCasa365 SFT expert bank and 900 demonstration-first observations per
condition, with 50 observations from each task. The lexical/syntactic templates
change command wording and structure while preserving task-critical slots;
the goal-state templates describe the same desired outcomes. All selectors
remain fixed during evaluation, and test task identities are used only to
compute metrics.

\paragraph{RoboCasa365 comparison.}
Table~\ref{tab:robocasa-selection-baselines} reports instruction robustness
and native-instruction closed-loop success. \policyweave leads on native
and lexical/syntactic instructions, whereas Semantic-NN leads on goal-state
rephrasings. The closed-loop results follow the native-instruction ranking.
\begin{table}[!ht]
  \centering
  \caption{RoboCasa365 SFT selection baselines (\%). Routing accuracy uses
  900 recorded contexts per condition; closed-loop success uses 50 episodes
  per task. Sem-NN denotes task-name-based semantic retrieval.}
  \label{tab:robocasa-selection-baselines}
  \setlength{\tabcolsep}{4pt}
  \renewcommand{\arraystretch}{1.08}
  \begin{tabular*}{\linewidth}{@{\extracolsep{\fill}}lrr>{\columncolor{policyweaveblue}[\tabcolsep][0pt]}r@{}}
    \toprule
    Condition & Trainable MLP & Sem-NN & \multicolumn{1}{r}{\policyweave} \\
    \midrule
    \multicolumn{4}{@{}l}{\textit{Top-1 routing accuracy}} \\
    Native & 94.33 & 80.56 & \textbf{99.44} \\
    Lexical/syntactic & 79.67 & 83.78 & \textbf{91.33} \\
    Goal-state & 67.78 & \textbf{82.56} & 81.78 \\
    \midrule
    \multicolumn{4}{@{}l}{\textit{Closed-loop success}} \\
    Native & 63.1 & 62.0 & \textbf{64.7} \\
    \bottomrule
  \end{tabular*}
\end{table}

On initial rollout observations, Top-1 accuracy is 92.33\% for the MLP,
81.44\% for Semantic-NN, and 98.67\% for \policyweave.

\paragraph{Observation timing.}
We evaluate the RoboCasa365 SFT then RL bank on recorded observations at
timesteps 0, 16, 32, and 48 from the same 50 demonstrations per task.
Instructions remain unchanged, and camera views and robot states are aligned
at each timestep. Across these 900 contexts per timestep, Top-1 accuracy
remains above 99\% (Table~\ref{tab:observation-timing}), indicating that
selection does not require the exact initial frame.
\begin{table}[!ht]
  \centering
  \caption{\policyweave expert identification from early-trajectory
  RoboCasa365 observations (900 contexts per timestep).}
  \label{tab:observation-timing}
  \setlength{\tabcolsep}{6pt}
  \renewcommand{\arraystretch}{1.05}
  \begin{tabular*}{\linewidth}{@{\extracolsep{\fill}}lrrrr@{}}
    \toprule
    Timestep & 0 & 16 & 32 & 48 \\
    \midrule
    Top-1 accuracy (\%) & 99.44 & 99.56 & 99.89 & 99.89 \\
    \bottomrule
  \end{tabular*}
\end{table}

\subsection{PolicyWeave Scorer Ablation}
\label{app:contextvote-scorer-ablation}
We compare scorer variants on 900 initial demonstration observations from
the RoboCasa365 SFT bank, with 50 per task and native instructions. A base
pass supplies context at eight cross-attention layers using a zero action
latent and timestep zero. The 66 configurations compare key ($\mathrm{K}$),
value ($\mathrm{V}$), and combined key-and-value ($\mathrm{K+V}$) update
responses, with raw or normalized scores across individual layers, the first
or last four layers, and all eight layers. Cross 0--7 index the cross-attention
layers in network order. Normalized scores follow
Equation~\eqref{eq:contextvote-score}; raw scores use its numerator.
The $\mathrm{K+V}$ variant adds the key and value scores within each layer
before aggregating across layers.

Table~\ref{tab:contextvote-scorer-full} shows that accuracy varies with depth.
Cross 3 gives the strongest single-layer $\mathrm{K}$ result. Although
$\mathrm{V}$ scores often perform better at individual layers, aggregating
all eight layers favors $\mathrm{K}$ over both $\mathrm{V}$ and $\mathrm{K+V}$,
with either raw or normalized scores.
\begin{table}[!ht]
  \centering
  \caption{Top-1 routing accuracy (\%) for all 66 scorer configurations on
  native instructions.}
  \label{tab:contextvote-scorer-full}
  \renewcommand{\arraystretch}{0.96}
  \setlength{\tabcolsep}{2.8pt}
  \begin{tabular*}{\textwidth}{@{\extracolsep{\fill}}lrrrrrr@{}}
    \toprule
    & \multicolumn{3}{c}{Normalized response} &
      \multicolumn{3}{c}{Raw response} \\
    \cmidrule(lr){2-4}\cmidrule(l){5-7}
    Scoring layers & $\mathrm{K}$ & $\mathrm{V}$ & $\mathrm{K+V}$ &
      $\mathrm{K}$ & $\mathrm{V}$ & $\mathrm{K+V}$ \\
    \midrule
    Cross 0 & 83.33 & 83.89 & 89.00 & 76.89 & 77.89 & 85.89 \\
    Cross 1 & 95.78 & 97.56 & 98.00 & 96.33 & 98.00 & 98.67 \\
    Cross 2 & 93.11 & 95.56 & 94.78 & 93.11 & 96.33 & 95.67 \\
    Cross 3 & \textbf{99.56} & 98.33 & 99.22 & \textbf{99.56} & 96.89 & \textbf{99.56} \\
    Cross 4 & 96.67 & \textbf{99.11} & 99.22 & 96.11 & 97.22 & 98.11 \\
    Cross 5 & 97.56 & 99.00 & \textbf{99.44} & 97.56 & \textbf{98.89} & 99.00 \\
    Cross 6 & 97.56 & 97.78 & 98.89 & 97.44 & 95.44 & 98.11 \\
    Cross 7 & 94.67 & 95.11 & 96.67 & 94.11 & 83.33 & 90.89 \\
    Early 4 & 98.89 & 97.56 & 98.56 & 99.00 & 98.11 & 99.33 \\
    Late 4  & 99.33 & 98.89 & 99.33 & 99.11 & 97.44 & 98.56 \\
    All 8   & 99.44 & 98.56 & 99.11 & 99.44 & 97.89 & 99.00 \\
    \bottomrule
  \end{tabular*}
\end{table}

Across the 18 multi-layer variants, adaptive selection gives 98.56--100\%
expert recall with mean $k$ between 1.006 and 1.042. Using all eight layers,
normalized $\mathrm{K}$ scores achieve 99.44\% recall at mean $k=1.006$,
while raw $\mathrm{K}$ scores achieve 99.78\% at $1.007$.

\subsection{Scaling the Expert Bank}
\label{app:contextvote-bank-scaling}
We vary bank size over $K\in\{1,2,4,8,12,18\}$ on NavigateKitchen,
PickPlaceSinkToCounter, CloseFridge, OpenCabinet, and TurnOnMicrowave,
using 50 initial demonstration observations per task with native instructions.
Each bank includes the target expert and adds distractors in descending
order of mean key-response score measured in a prior routing study.
Thus, $K=2$ includes the strongest measured distractor. Task identities are
used to construct and evaluate the subbanks, not as routing inputs.

\begin{table}[!ht]
  \centering
  \caption{PolicyWeave statistics as the available expert bank grows. Each
  bank size is evaluated on 250 native-instruction observations. Routing time
  reports p50 / p95 latency for a batch of five with the model and expert bank
  resident on the GPU.}
  \label{tab:contextvote-bank-scaling}
  \renewcommand{\arraystretch}{0.98}
  \setlength{\tabcolsep}{3.5pt}
  \begin{tabular*}{\textwidth}{@{\extracolsep{\fill}}rrrrrr@{}}
    \toprule
    Experts & Top-1 (\%) $\uparrow$ & Expert recall (\%) $\uparrow$ & Mean $k$ $\downarrow$
    & \shortstack{Storage\\($10^6$ bytes)} & Route p50 / p95 (ms) \\
    \midrule
     1 & 100.00 & 100.00 & 1.000 &  19.17 & 80.67 / 146.15 \\
     2 &  98.00 &  98.00 & 1.020 &  38.35 & 80.22 / 143.20 \\
     4 &  98.00 &  98.00 & 1.020 &  76.70 & 79.89 / 140.64 \\
     8 &  98.00 &  98.00 & 1.020 & 153.39 & 84.73 / 128.61 \\
    12 &  98.00 &  98.00 & 1.020 & 230.09 & 83.45 / 142.28 \\
    18 &  98.00 &  98.00 & 1.020 & 345.14 & 90.61 / 139.86 \\
    \bottomrule
  \end{tabular*}
\end{table}

After the strongest distractor is introduced at $K=2$, Top-1 accuracy and
expert recall remain 98.00\% through $K=18$. Exact adapter storage grows linearly,
while routing is performed once when the context-conditioned merged update is
formed.

\section{Real-Robot Experiments}
\label{app:real-robot}
\subsection{Setup and Training}
\paragraph{Platform and tasks.}
The policy controls one AgileX Piper arm with six joints and a parallel
gripper. Observations contain two RGB camera views, the instruction, and
seven-dimensional joint-and-gripper state. Actions specify six absolute
joint targets in radians and gripper opening in meters. Demonstrations are
recorded at 30\,Hz. The four tasks are \textit{Upright bottle}, which places a
bottle lying on the table upright; \textit{Stack blocks}, which stacks two
blocks; \textit{Press button}, which activates a light using the black button;
and \textit{Place cup}, which places an empty cup on the white coaster.

\paragraph{Shared adaptation and task experts.}
Each task has 51 demonstrations. Starting from pretrained GR00T N1.5, we
adapt the action model and its input-output projectors jointly on one
demonstration per task for 3,000 steps, keeping the VLM frozen. This produces
the Piper shared base. State and action normalization use pooled 1st and 99th
percentiles from all 204 demonstrations. We then preserve this normalization
and the adapted interface while training one rank-8, scale-16 LoRA expert on
the remaining 50 demonstrations of each task for 10,000 steps. LoRA updates
cover hidden action-transformer linear layers, excluding the input-output
projectors. Table~\ref{tab:piper-training} summarizes the training settings.

\paragraph{Co-training.}
Both co-training baselines jointly use all 51 demonstrations per task and
train for up to 40,000 steps. GR00T co-training uses the same adapted base,
frozen action interface, LoRA configuration, and optimizer settings as the
single-task experts; only the task mixture and maximum training steps change.
For $\pi_{0.5}$, we use uniform task sampling, freeze the VLM, train rank-8,
scale-16 LoRA updates
in action attention and feed-forward layers, and update the action
input-output projections, timestep embedding, and adaptive normalization.
Both the GR00T and $\pi_{0.5}$ training
pipelines use AdamW with weight decay $10^{-5}$, 5\% learning-rate warmup
followed by cosine decay, and bfloat16 computation.
\begin{table}[!ht]
  \centering
  \caption{Piper training configurations. Batch sizes are effective totals;
  the expert configuration applies independently to each of the four tasks.}
  \label{tab:piper-training}
  \small
  \setlength{\tabcolsep}{4pt}
  \renewcommand{\arraystretch}{1.12}
  \begin{tabularx}{\linewidth}{@{}l*{4}{>{\centering\arraybackslash}X}@{}}
    \toprule
    Setting & Shared GR00T adaptation & \policyweave experts & GR00T co-training & $\pi_{0.5}$ co-training \\
    \midrule
    Demonstrations & 1 per task & 50 per expert & 51 per task & 51 per task \\
    Maximum steps & 3,000 & 10,000 & 40,000 & 40,000 \\
    Batch size & 128 & 256 & 256 & 256 \\
    Gradient accumulation & 8 & 1 & 1 & 1 \\
    Peak learning rate & $3\times10^{-5}$ & $10^{-4}$ & $10^{-4}$ & $10^{-4}$ \\
    LoRA rank / scale & -- & 8 / 16 & 8 / 16 & 8 / 16 \\
    VLM & Frozen & Frozen & Frozen & Frozen \\
    \bottomrule
  \end{tabularx}
\end{table}

\paragraph{Inference and evaluation.}
Both model families predict 16-step action chunks. GR00T uses four flow
integration steps and $\pi_{0.5}$ uses ten. \policyweave uses the same eight
cross-attention key projections and stability-adaptive selection rule as in
the simulation experiments, fixing the merged update for each episode.
Each method is evaluated over 20 rollouts per task. We report the fraction
of successful rollouts for each task and the unweighted mean across the four tasks.
\subsection{Real-Robot Demonstrations}
\label{app:real-robot-demos}
\begin{figure}[p]
  \centering
  \includegraphics[width=\linewidth]{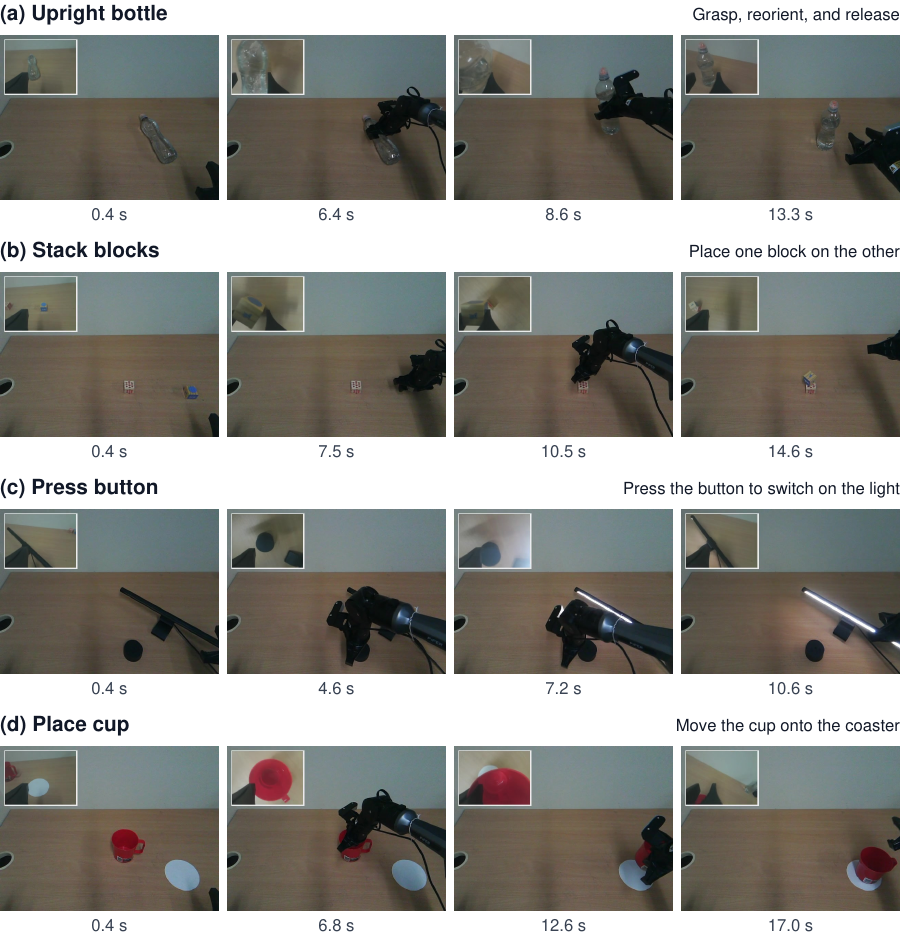}
  \caption{Successful \policyweave rollouts on the Piper arm. Each row follows
  one video from left to right; timestamps indicate elapsed video time.
  Insets show the wrist-camera view.}
  \label{fig:piper-success-sequences}
\end{figure}

Figure~\ref{fig:piper-success-sequences} presents one successful \policyweave
rollout for each Piper task. The sequences show bottle reorientation, block
stacking, button activation, and cup placement from the initial scene to task
completion. Full videos are included in the supplementary material.

\section{Limitations}
\label{sec:limitations}
Our experiments use GR00T N1.5, with real-world evaluation on a Piper arm;
other VLA architectures and broader deployment settings remain to be studied.
\policyweave merges experts derived from a common base policy with a shared
embodiment and action interface. Extending it across embodiments or to tasks
requiring additional action channels would require further interface alignment.
The merged parameters remain fixed throughout each episode, so the method does
not revise expert selection as new observations arrive. Tasks whose identity
becomes clear only through interaction, or whose goals change during execution,
would require a mechanism for updating the selection and merged policy online.
Our real-world experiments evaluate \policyweave, while static
merging is evaluated only in simulation. Extending static merging to physical
robots calls for reliable coefficient selection and safety validation.
Developing static merging methods better suited to VLA control and real-world
deployment remains an important direction.

\end{document}